\documentclass[conference]{IEEEtran}
\IEEEoverridecommandlockouts
\usepackage{cite}
\usepackage{amsmath,amssymb,amsfonts}
\usepackage{algorithmic}
\usepackage{graphicx}
\usepackage{textcomp}
\usepackage{xcolor}

\usepackage{multirow}
\definecolor{lightgray}{gray}{.9}
\usepackage{colortbl}
\usepackage{booktabs}
\usepackage{algorithm}
\usepackage{algorithmic}

\def\BibTeX{{\rm B\kern-.05em{\sc i\kern-.025em b}\kern-.08em
    T\kern-.1667em\lower.7ex\hbox{E}\kern-.125emX}}
\begin{document}

\title{Adaptive Operator Learning for Multiphysics PDEs via Shapley-Guided Field Bias Mitigation
\\
\thanks{*These authors contributed equally to this work.}
\thanks{†Zhenzhong Wang is the corresponding authors.}
}

\author{\IEEEauthorblockN{\qquad \quad \;\:\ \ \ \ Chunhe Liu*\ \ \qquad \quad \ \;\:\ }
\IEEEauthorblockA{\textit{The University of Hong Kong \&} \\
\textit{Xiamen University}\\
Hong Kong, China \\
u3637314@connect.hku.hk}
\and
\IEEEauthorblockN{\qquad \quad \;\:\ \ \ \ \ Jihang Wu*\ \ \qquad \quad \ \;\:\ }
\IEEEauthorblockA{\textit{The Hong Kong Polytechnic University} \\
Hong Kong, China \\
mieux052@gmail.com}
\and
\IEEEauthorblockN{\qquad \quad\;\:\ \ \ \ Shujia Ding*\ \qquad \quad \ \;\:\ }
\IEEEauthorblockA{\textit{Xiamen University} \\
Xiamen, China \\
37220222203577@stu.xmu.edu.cn}
\and
\IEEEauthorblockN{\qquad \quad\;\:\ \ \ \ \ \ Fengye Chen\ \ \qquad \quad \ \;\:\ }
\IEEEauthorblockA{\textit{Xiamen University} \\
Xiamen, China \\
22920252203283@stu.xmu.edu.cn}
\and
\IEEEauthorblockN{\qquad \quad\;\:\ \ Zhenzhong Wang†\qquad \quad \ \;\:\ }
\IEEEauthorblockA{\textit{Xiamen University} \\
Xiamen, China \\
zhenzhongwang@xmu.edu.cn}
\and
\IEEEauthorblockN{\qquad \quad\;\:\ \ \ \ \ Min Jiang\ \ \qquad \quad \ \;\:\ }
\IEEEauthorblockA{\textit{Xiamen University} \\
Xiamen, China \\
minjiang@xmu.edu.cn}
}

\maketitle

\begin{abstract}
Multiphysics simulation is indispensable for modeling complex systems governed by coupled Partial Differential Equations (PDEs). However, traditional numerical methods face prohibitive computational costs. To accelerate this process, Neural Operators (NOs) have emerged. While promising, when applying existing NOs to coupled PDEs, they frequently lead to a persistent challenge: We identify it as {\em field dominance bias}. We find that this bias occurs because the field with higher energy dominates the optimization landscape. This results in the underlearning and poor prediction accuracy of the low-energy fields. To mitigate this issue, we propose the Adaptive Gradient Scaling Neural Operator (AGNO). AGNO achieves balanced field modeling by disentangling the backpropagated gradients from each coupled field. We pioneer the use of the Shapley Value to quantify each field's contribution to the total loss, thereby diagnosing which fields are dominant or under-learned. This quantification directs an adaptive gradient scaling mechanism: gradients are amplified for under-learned fields and reduced for dominant ones. This approach effectively alleviates field dominance bias, leading to an up $\sim$18\% improvement in overall prediction accuracy and demonstrating superior generalization in both low-data and noisy environments compared to existing baselines.
\end{abstract}


\section{Introduction}

Multiphysics simulation is a cornerstone of computational science and engineering, proving indispensable for modeling complex systems and physical phenomena. At its core, multiphysics involves the numerical solution of coupled systems of Partial Differential Equations (PDEs), where each equation governs a distinct physical process—such as fluid flow, structural mechanics, heat transfer, or electromagnetics~\cite{Wong2025EvolutionaryOO,vinuesa2023transformative,ni2022soft,keyes2013multiphysics}. This capability allows engineers and scientists to describe many real-world systems, ranging from predicting the thermo-mechanical stress in aerospace components and optimizing the electrochemical response in battery design to modeling the fluid-structure interaction of biological systems~\cite{liu2025difffno,fno,li2023fourier,bonev2023spherical,lu2021learning}.

Traditionally, solving PDEs relies on numerical methods such as the finite difference, finite element, etc \cite{fem,fv,Spectral}. These approaches typically necessitate fine spatial discretization to capture physical phenomena, leading to prohibitively high computational costs. To address these limitations, recent research has focused on deep learning-based alternatives, notably through Neural Operators ($\text{NO}$s), which model mappings between infinite-dimensional function spaces \cite{chen1,chen2,fno,liu2025difffno,bonev2023spherical,lu2021learning}. The concept of NOs was initially realized with DeepONet \cite{lu2021learning, dyna}, which employs two interconnected deep neural networks to process input functions and query coordinates. This design led to various subsequent variants \cite{lu2021learning,xu2023transfer,mei2024fully,wang2021learning,qiu2024derivative}, such as $\text{MIONet}$ \cite{mion}, an extension designed to accommodate multiple physical fields. Other methods implement neural operators by operating in the spectral space. The representative Fourier Neural Operator ($\text{FNO}$) \cite{fno} utilizes the Fourier transform to approximate the kernel integration. Subsequent extensions have built upon $\text{FNO}$ to address complexities \cite{liu2025difffno,li2023fourier, bonev2023spherical,li2024multi,wang2024transfer}. For example, $\text{CoDA-NO}$ \cite{coda-no} extends the $\text{FNO}$ by treating physical variables as infinite-dimensional tokens, enabling the efficient handling of PDEs.

\begin{figure}[t]
\begin{center}
\includegraphics[width=0.47\textwidth]{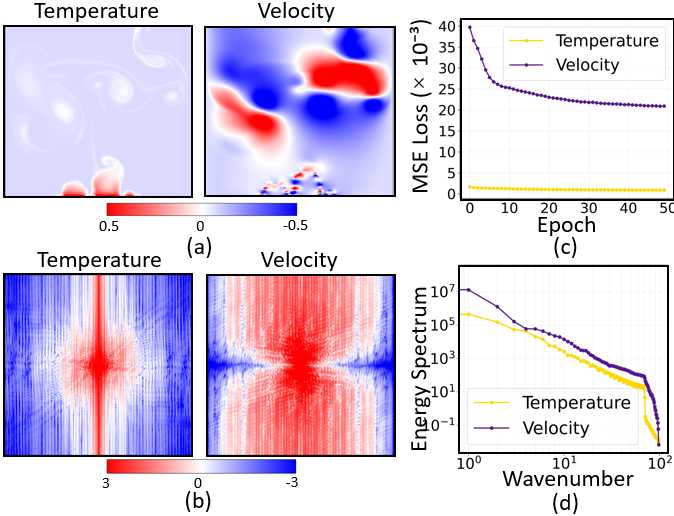}
\end{center}
\vspace{-0.4cm}
\caption{(a). Temperature and velocity fields in a multiphase flow system. (b). Energy spectra of temperature and velocity fields. (c). Validation loss comparison between temperature and velocity fields. (d). Radially averaged energy spectra of temperature and velocity fields.}
\label{fig:intro}
\vspace{-0.4cm}
\end{figure}

Despite recent progress, we observe that in multi-physics settings, existing approaches induce an imbalance learning across multiple physical fields, causing under-learning of certain fields \cite{hae26,clink,liu26}. We formalize this overlooked issue as {\em field dominance bias}.
To illustrate this, we consider a representative multiphysics application, such as multiphase flow. As visualized in Fig. \ref{fig:intro} (a-b), the temperature field and the velocity field exhibit markedly different distributions of high-frequency features and energy. The velocity field contains more numerous and localized sharp discontinuities. Our Fourier transform analysis, shown in Fig. \ref{fig:intro} (d), confirms this observation: the velocity field possesses significantly higher energy across various wavenumbers compared to the temperature field. This energy disparity creates a challenging optimization landscape. The high-energy, high-frequency velocity field naturally generates a larger initial loss and consequently larger gradients during backpropagation (see Fig. \ref{fig:intro} (c)). This large gradient magnitude can dominate the learning process, resulting in poor prediction for the low-energy field, i.e., the temperature field, particularly inhibiting the learning of its subtle high-frequency features.

It is important to distinguish the proposed field dominance bias from the well-known spectral bias~\cite{kang2025pig,LIU2024112944,wangg2026work,ppc}. Spectral bias describes the intrinsic tendency to learn low-frequency components of a single function before fitting its high-frequency details, and is therefore a property of function approximation within one field. In contrast, field dominance bias arises in multi-field learning scenarios, where different physical fields exhibit inherently different energy and frequency distributions. The issue is not that high-frequency components are harder to learn, but that fields with larger overall energy and richer high-frequency content produce disproportionately larger gradients during optimization, thereby dominating parameter updates. As a result, other fields—despite possibly containing important fine-scale structures—are under-learned due to cross-field competition rather than intra-field frequency preference. This distinction highlights that field dominance bias is an optimization imbalance across coupled fields, rather than a frequency preference within a single field.

To mitigate this bias, this work proposes an Adaptive Gradient Scaling Neural Operator (AGNO). The core design lies in gradient disentanglement during backpropagation, which balances the learning pace across diverse physical fields. Existing approaches such as Dual-Balancing \cite{Dual-Balancing} (based on GradNorm \cite{grad}) and CONFIG \cite{ConFIG} only align gradient magnitudes or directions without decoupling physical gradients. By contrast, Shapley values can accurately quantify the marginal loss contribution of each physical field, thereby effectively decoupling coupled physical systems and alleviating the learning imbalance in multiphysics tasks. 

Specifically, we introduce the Shapley value to quantify the contribution of the gradient of different physical fields to the overall prediction. Based on each field's contribution, we apply a gradient scaling mechanism to control the gradient magnitudes: Gradients are amplified for the under-learned fields, while gradients are reduced for the dominant fields. This adaptive process alleviates field dominance bias, ultimately leading to an improvement in overall prediction accuracy for all coupled physical fields. Our contributions can be summarized as follows:

\begin{itemize}
    \item We first mitigate the field dominance bias in multiphysics PDEs. Particularly, we disentangle the backpropagated gradients from each coupled physical field. This disentanglement allows for the adaptive and independent manipulation of the learning signals for each field, ensuring a balanced optimization trajectory for all fields.

    \item We pioneer the use of the Shapley value in the context of multiphysics NOs. The Shapley value quantifies the contribution of each individual physical field to the prediction, serving as a diagnostic tool to measure which fields are under-learned or dominant, directly identifying the source of the optimization imbalance.

    \item Experimental results demonstrate that AGNO significantly outperforms existing baselines, achieving an improvement of approximately 18\% in prediction accuracy. Moreover, AGNO shows strong generalization capability in low-data super-resolution scenarios as well as in tasks involving noisy inputs.
\end{itemize}

\begin{figure*}[h]
\begin{center}
\includegraphics[width=1\textwidth]{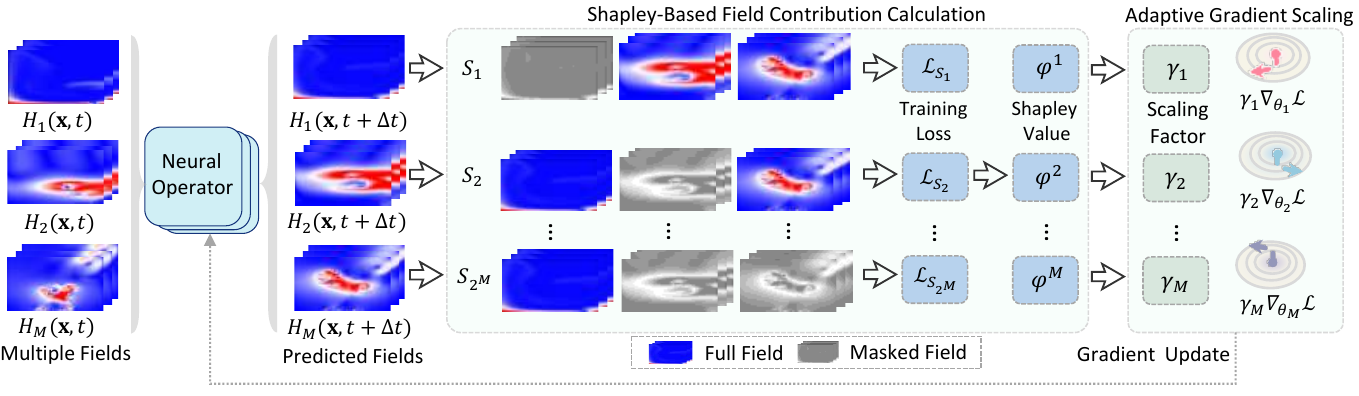}
\end{center}
\vspace{-0.4cm}
\caption{The proposed AGNO devises a field contribution calculation mechanism to calculate the contribution of each physical field to the prediction. The gradient scaling mechanism calculates the gradient scaling factor for each physical field and scales the gradient of each physical field during backpropagation.}
\label{fig:framework}
\vspace{-0.4cm}
\end{figure*}

\section{Methodology}

\subsection{The Shapley Value}
The Shapley value ($\phi$) is a foundational concept in cooperative game theory, first introduced by Lloyd Shapley~\cite{roth1988introduction,algaba2019handbook}. It provides a principled and unique solution for fairly distributing the total value (or "payoff") generated by a coalition of players among its individual members, based on their marginal contributions.

A coalitional game is formally defined by a set of players $N$ and a characteristic function $v: 2^N \to \mathbb{R}$, where $v(S)$ represents the total value that a coalition $S \subseteq N$ can achieve by cooperating. The Shapley value $\phi_i(v)$ for a player $i \in N$ is defined as their average marginal contribution across all possible ordering (permutations) of the players. Mathematically, $\phi_i(v)$ is expressed as:
\begin{equation}
    \phi_i(v) = \sum_{S \subseteq N \setminus \{i\}} \frac{|S|! (|N| - |S| - 1)!}{|N|!} [v(S \cup \{i\}) - v(S)],
\end{equation}
where the term $v(S \cup \{i\}) - v(S)$ is the marginal contribution of player $i$ to the existing coalition $S$. The formula essentially weights this marginal contribution by the probability of encountering the coalition $S$ in a random ordering of all players.

In the field of machine learning, the Shapley value has been powerfully repurposed for feature attribution, a concept popularized by the $\mathrm{SHAP}$ (SHapley Additive exPlanations) framework \cite{rawal2021recent,zhao2025shapley}. This application redefines the core elements of the cooperative game: the input features $\mathcal{F}$ of the machine learning model act as the set of players ($N$), and any subset of these features forms a coalition ($S$). The characteristic function ($v$) is then defined by the model's output when only the features in $S$ are available, typically measured as the expected model output conditioned on the known features: $v(S) = \mathbb{E}[f(x) \mid x_S]$. Consequently, the resulting Shapley value, $\phi_i(v)$, quantifies the contribution of the $i$-th feature to the model's specific prediction, measured against a predefined baseline expectation. This formulation provides a locally accurate and model-agnostic methodology for interpreting complex black-box outputs, crucially ensuring that all feature contributions are fairly allocated according to the rigorous, axiomatic principles derived from cooperative game theory.

\subsection{Framework Overview}

As depicted in Fig. \ref{fig:framework}, we adopt FNO\footnote{The performance of NO based on Transformer architecture is explored in Table \ref{xr}.} as the neural operator to encode the solution field at the historical timestep, $(H_1(\cdot,t), \ldots, H_M(\cdot,t))$, and predict the evolution of the multi-physics dynamics, $H(\cdot, t+\Delta t)$. Following this prediction, the model computes the loss for all output subsets separately, utilizing the Shapley value as a diagnostic tool to quantify the influence of each physical field on the overall prediction output. This contribution score (i.e., the Shapley value) is then translated into an optimization signal via the gradient scaling mechanism. The final gradient for each field is scaled by this factor, which allows us to independently modulate the learning rate of each field. This dynamic control over the learning extent for individual fields facilitates a balanced optimization trajectory, mitigating the observed field dominance bias. The details of the two key modules, Shapley-Based Field Contribution Calculation and Adaptive Gradient Scaling, are presented in what follows.

\subsection{Shapley-Based Field Contribution Calculation}

One primary challenge is to quantify how much each individual physical field contributes to the overall prediction error, thereby identifying which fields are being under-learned or dominant. Traditional loss functions fail to provide this attribution, as the total loss is a coupled scalar value. To overcome this, we leverage the Shapley value ($\phi$), which offers a fair allocation of the total loss among the input features (the physical fields).

Let $\mathbf{x} = \{\mathbf{x}_{1}, \mathbf{x}_{2}, \dots, \mathbf{x}_M\}$ is the multiphysics input, the total number of physical fields is $M$. Let $f(\mathbf{x})$ denote the output prediction of the NO given the input $\mathbf{x}$. The main objective of model optimization is to maximize the model's prediction accuracy relative to multiphysics data. To this end, the training objective is set to minimize the mean squared error (MSE), which quantifies the L2 norm distance between the model's predictions and the target values.
$
\mathcal{L}(\mathbf{x}) = \frac{1}{|F|}\sum_{\mathbf{x} \in F}(\hat{f}(\mathbf{x})-f(\mathbf{x}))^2
$, where $F$ denotes all points in the physical field, $\hat{f}(\mathbf{x})$ represents the model's output at location $\mathbf{x}$. Our goal is to determine the contribution of a specific field $m \in M$, and to allocate the model's predictions loss to each physics field. Denote the Shapley value of field $m$ as $\phi^{m}(\mathbf{x})$, which satisfy:
$
\sum_{m\ \in \ M}\phi^{m}(\mathbf{x}) \equiv \mathcal{L}(\mathbf{x})
$. The Shapley value $\phi^{m}(\mathbf{x})$ is defined as the weighted average of the {marginal contribution} of field $m$ across all possible subsets (coalitions) $S$ of the other fields $M/\{m\}$:

\begin{equation}
\phi^{m}(x)=\sum_{S\ \subseteq M/\{m\}}\frac{\left|S\right|!\left(k\ -\ \left|S\right|-\ 1\right)!}{k!}V_{m\left(S;\ \phi\right)}.\label{sha}
\end{equation}
Here, the term $k$ represents the total number of the fields. The factor $\frac{|S|!(k - |S| - 1)!}{k!}$ represents the normalized weight (or frequency) of the subset $S$ appearing before field $m$ in all possible orderings. The marginal contribution of physical field $m$ to the subset of physical fields $S$, defined as $V_{m(S;\ \phi)}$ is calculated as:
\begin{equation}
V_{m(S;\ \phi)}\ =\ \mathcal{L}(S\ \cup\{m\})\ -\ \mathcal{L}(S),
\end{equation}
where the marginal effect $V_{m(S;\ \phi)}$ measures the change in the NO's loss when the field $m$ is added to the existing subset of fields $S$. By averaging this effect over all possible coalitions $S$, the Shapley value $\phi^{m}(\mathbf{x})$ provides an {equitable measure of influence} for field $m$ that accounts for all coupling effects.

To compute the Shapley value, the NO must provide an output $f(S)$ for every subset $S$ of the input fields to compute the loss $\mathcal{L}(S)$ of each subset. 
For instance, consider a system with two fields, $M = \{T, V\}$. The Shapley value for the temperature field, defined as $\phi^T$, is given by:
\begin{equation}
\phi^T(\mathbf{x})=\frac{1}{2}\left[\mathcal{L}\left(\{T,V\}\right)-\mathcal{L}\left(V\right)+\mathcal{L}\left(T\right)\right].
\end{equation}
The calculated Shapley value $\phi^{m}(\mathbf{x})$ is a tensor with the same spatial dimension as the physical field data, allowing us to assess contribution at every point in the domain. The greater the contribution of the physical field to the loss, the more it is to be ignored by NOs in the field equilibrium deviation.

\subsection{Adaptive Gradient Scaling}

Having established a measure of contribution for each field using the Shapley value $\phi^m(\mathbf{x})$, we now introduce the adaptive gradient scaling mechanism. The primary intuition is to treat the contribution of each field as a feedback signal to modulate the learning rate, suppressing the overly dominant fields and boosting the under-learned ones.

The Shapley value guarantees that the sum of all field contributions equals the total system loss. Intuitively, if a field is accurately learned, its contribution $\phi^m(\mathbf{x})$ should close to 0. Consequently, a smaller $\phi^m(\mathbf{x})$ signifies that the field is better approximated and therefore more contributive to the final correct output. Since our goal is to enhance the gradient for fields with larger loss, a smaller $\phi^m(\mathbf{x})$ thus corresponds to a better-learned, more contributive field, which will subsequently receive gradient suppression.

To ensure stability throughout long-term training, we mitigate rapid fluctuations in the instant MSE-based contribution by applying an exponential moving average (EMA). The EMA contribution value $c_i^t$ retains sensitivity to recent performance while smoothing the learning signal:
\begin{equation}
c_m^t = \begin{cases}
 0.8\cdot c_m^{t-1} + 0.2\cdot\phi^m(\mathbf{x}), &t>1\\
 \phi^m(\mathbf{x}), &t=1 \label{ema}
\end{cases}
\end{equation}

We aim for the gradient scaling factor $\gamma_m^t$ to modulate the learning intensity relative to the average field performance. Specifically, the factor $\gamma_m^t$ should be designed such that: If $c_m^t$ is low (well-learned/dominant), $\gamma_m^t < 1$ to \emph{suppress} the field's gradient magnitude. If $c_m^t$ is high (under-learned), $\gamma_m^t > 1$ to \emph{amplify} the field's gradient magnitude. The reference point for balance is the average contribution $\overline{c}^t = \frac{1}{k}\sum_{k=1}^{n}c_k^t$. We use the exponential function of the difference $(c_m^t - \overline{c}^t)$ to calculate the scaling factor $\gamma_m^t$:
\begin{equation}
\gamma_m^t=\exp{\left(c_m^t - \overline{c}^t\right)}.\label{exp}
\end{equation}
This exponential form elegantly satisfies the required conditions: (1) It ensures $\gamma_m^t > 0$, preserving the gradient direction. (2) If $c_m^t < \overline{c}^t$ (dominant field), the exponent is negative, resulting in $\gamma_m^t < 1$ (suppression). (3) If $c_m^t > \overline{c}^t$ (under-learned field), the exponent is positive, resulting in $\gamma_m^t > 1$ (amplification).

The FNO structure includes Fourier layers that do not change the channel dimensions and projection layers that adjust the channel dimensions of the corresponding physical fields. The architecture of Fourier layers naturally separates the gradient corresponding to each input physics field during propagation. In the projection layer that adjusts the channel dimension, we use grouped $1 \times 1 $ convolution to ensure that the projected channel dimension still corresponds to the original input physics field.
$
\text{X}_{proj}=\oplus_{m}\text{Conv}(W_m,\ x_m)
$, where $W_m$ is a convolution kernel with a size of $1 \times 1$, $\oplus_{m}$ represents concatenation along the channel dimension. 

This grouped convolutional projection method can map $M$ groups channels after projection to the $M$ input physical fields. After the Fourier layer, we create a gradient hook for each group of channels, so that the gradient scaling factor can be mapped to the gradient of the physical field during backpropagation.

Finally, we apply this scaling factor to the backpropagated gradient during the parameter update. Let $\Theta$ denote the set of all learnable parameters, $\eta$ be the learning rate, and $\mathcal{L}$ be the total MSE loss. The gradient update rule is modified to incorporate the adaptive scaling:
\begin{equation}
\Theta^t \leftarrow \Theta^t - \eta \cdot \sum_{m=1}^{M} \left( \gamma_m^t \cdot \frac{\partial \mathcal{L}}{\partial f^m} \cdot \frac{\partial f^m}{\partial \Theta^t} \right),\label{af}
\end{equation}
where $f^m$ is the output of field $m$, and $\frac{\partial \mathcal{L}}{\partial f^m}$ and $\frac{\partial f^m}{\partial \mathbf{W}^t}$ are standard components of the chain rule. 

By adaptively multiplying the field-specific gradient component $\left( \frac{\partial \mathcal{L}}{\partial f^m} \cdot \frac{\partial f^m}{\partial \mathbf{W}^t} \right)$ by the factor $\gamma_m^t$, we achieve dynamic control over the learning speed of each physical field, effectively mitigating the spectral asymmetry induced bias.

\begin{table*}[h]\scriptsize
\centering
\resizebox{1\linewidth}{!}{
\begin{tabular}{ll||ccc|ccc|ccc|ccc|cc}
\specialrule{\heavyrulewidth}{0pt}{0pt}
 &  & \multicolumn{3}{c}{\textbf{Subcooled}} & \multicolumn{3}{c}{\textbf{Gravity}} & \multicolumn{3}{c}{\textbf{Saturated}} & \multicolumn{3}{c}{\textbf{Single Bubble}} & \multicolumn{2}{c}{\textbf{Drug Delivery}}\\
\cline{3-16}
\multirow{-2}{*}{\textbf{Model}} & \multirow{-2}{*}{\textbf{Metric}} & $\mathbf{T}$ & $\mathbf{V_x}$ & $\mathbf{V_y}$ & $\mathbf{T}$ & $\mathbf{V_x}$ & $\mathbf{V_y}$ & $\mathbf{T}$ & $\mathbf{V_x}$ & $\mathbf{V_y}$ & $\mathbf{T}$ & $\mathbf{V_x}$ & $\mathbf{V_y}$ & $\mathbf{C}$ & $\mathbf{P}$\\
\hline \hline
\multirow{4}{*}{MIONet} & RMSE & 0.048 & 0.231 & 0.221 & 0.139 & 0.581 & 0.448 & 0.054 & 0.518 & 0.549 & 0.028 & 0.103 & 0.098 & 0.045 & 0.017\\
& FRMSE low & 0.248 & 1.423 & 1.218 & 0.804 & 4.388 & 1.282 & 0.295 & 4.762 & 5.085 & 0.131 & 0.278 & 0.302 & -- & -- \\
& FRMSE mid & 0.271 & 1.125 & 1.084 & 0.758 & 3.369 & 2.890 & 0.372 & 3.458 & 3.891 & 0.118 & 0.291 & 0.338 & -- & -- \\
& FRMSE high & 0.045 & 0.218 & 0.187 & 0.133 & 0.541 & 0.431 & 0.051 & 0.442 & 0.488 & 0.023 & 0.079 & 0.081 & -- & -- \\
\hline
\multirow{4}{*}{U-Net} & RMSE & 0.036 & 0.185 & 0.165 & 0.126 & 0.792 & 0.425 & 0.046 & 0.408 & 0.409 & 0.032 & 0.064 & 0.071 & 0.063 & 0.162\\
& FRMSE low & 0.152 & 0.831 & 0.968 & 0.807 & 4.473 & 1.279 & 0.168 & 2.815 & 2.936 & 0.312 & 0.528 & 0.842 & -- & -- \\
& FRMSE mid & 0.219 & 0.972 & 0.928 & 0.696 & 3.118 & 2.282 & 0.315 & 3.295 & 3.517 & 0.285 & 0.594 & 0.563 & -- & -- \\
& FRMSE high & 0.039 & 0.174 & 0.142 & 0.120 & 0.504 & 0.392 & 0.044 & 0.368 & 0.346 & 0.027 & 0.055 & 0.062 & -- & -- \\
\hline
\multirow{4}{*}{CODA-NO} & RMSE & 0.037 & 0.201 & 0.178 & 0.117 & 0.534 & 0.437 & 0.048 & 0.445 & 0.442 & -- & -- & -- & 0.047 & 0.028\\
& FRMSE low & 0.175 & 1.128 & 1.135 & 0.484 & 3.630 & 1.342 & 0.213 & 3.725 & 3.768 & -- & -- & -- & -- & -- \\
& FRMSE mid & 0.208 & 1.042 & 0.985 & 0.676 & 3.430 & 3.036 & 0.341 & 3.486 & 3.542 & -- & -- & -- & -- & -- \\
& FRMSE high & 0.036 & 0.192 & 0.157 & \textbf{0.112} & 0.495 & 0.416 & 0.047 & 0.405 & 0.384 & -- & -- & -- & -- & -- \\
\hline
\multirow{4}{*}{GNOT} & RMSE & 0.038 & 0.211 & 0.217 & 0.118 & 0.661 & 0.488 & 0.051 & 0.475 & 0.491 & 0.010 & 0.069 & 0.070 & 0.055 & 0.020\\
& FRMSE low & 0.185 & 1.285 & 1.698 & 0.696 & 5.781 & 3.543 & 0.271 & 3.985 & 4.428 & 0.129 & 0.728 & 0.967 & -- & -- \\
& FRMSE mid & 0.221 & 0.988 & 1.023 & 0.664 & 3.630 & 3.461 & 0.362 & 3.685 & 3.792 & 0.078 & 0.741 & 0.512 & -- & -- \\
& FRMSE high & 0.037 & 0.183 & 0.195 & 0.113 & 0.611 & 0.449 & 0.048 & 0.427 & 0.419 & 0.009 & 0.063 & 0.057 & -- & -- \\
\hline
\multirow{4}{*}{Transolver} & RMSE & 0.038 & 0.188 & 0.162 & 0.121 & 0.539 & 0.435 & 0.049 & 0.464 & 0.464 & 0.012 & 0.074 & 0.082 & 0.120 & 0.097\\
& FRMSE low & 0.216 & 1.002 & 0.957 & 0.529 & 3.244 & 1.824 & 0.264 & 3.896 & 3.961 & 0.153 & 0.992 & 1.289 & -- & -- \\
& FRMSE mid & 0.209 & 0.970 & 0.887 & 0.670 & 3.414 & 2.812 & 0.354 & 3.535 & 3.593 & 0.101 & 0.725 & 0.610 & -- & -- \\
& FRMSE high & 0.035 & 0.176 & 0.149 & 0.117 & 0.507 & 0.417 & 0.046 & 0.414 & 0.413 & 0.010 & 0.064 & 0.070 & -- & -- \\
\hline
\multirow{4}{*}{M2NO} & RMSE & 0.034 & 0.167 & 0.146 & 0.118 & 0.498 & \textbf{0.390} & 0.048 & 0.355 & 0.367 & 0.050 & 0.123 & 0.160 & 0.048 & 0.030\\
& FRMSE low & 0.118 & 0.655 & 0.696 & 0.451 & 3.048 & 1.352 & 0.223 & 2.500 & 2.718 & 0.197 & 0.568 & 0.717 & -- & -- \\
& FRMSE mid & 0.195 & 0.887 & 0.783 & 0.622 & 2.948 & 2.187 & 0.349 & 3.820 & 3.878 & 0.182 & 0.391 & 0.701 & -- & -- \\
& FRMSE high & 0.032 & 0.159 & 0.129 & 0.114 & 0.461 & 0.400 & 0.045 & 0.364 & 0.374 & 0.045 & 0.107 & 0.121 & -- & -- \\
\hline
\multirow{4}{*}{Dual-Balancing$_{\text{FNO}}$} & RMSE & 0.046 & 0.532 & 0.345 & 0.120 & \textbf{0.188} & 0.533 & 0.054 & 0.572 & 0.391 & 0.043 & 0.034 & 0.048 & 0.052 & 0.028\\
& FRMSE low & 0.334 & 11.402 & 7.121 & 0.729 & 3.055 & 15.037 & 0.321 & 16.313 & 7.521 & 0.801 & 0.710 & 0.911 & -- & -- \\
& FRMSE mid & 0.272 & \textbf{0.143} & \textbf{0.468} & 0.665 & \textbf{1.754} & \textbf{0.335} & 0.413 & \textbf{0.348} & \textbf{0.771} & 0.316 & 0.250 & 0.252 & -- & -- \\
& FRMSE high & 0.041 & \textbf{0.031} & \textbf{0.095} & 0.114 & \textbf{0.135} & \textbf{0.045} & 0.051 & \textbf{0.054} & \textbf{0.123} & 0.036 & 0.028 & 0.040 & -- & -- \\
\hline
\multirow{4}{*}{ConFIG$_{\text{FNO}}$} & RMSE & 0.037 & 0.182 & 0.145 & -- & -- & -- & -- & -- & -- & 0.014 & 0.055 & 0.056 & 0.081 & 0.036\\
& FRMSE low & 0.159 & 0.667 & 0.732 & -- & -- & -- & -- & -- & -- & 0.185 & 0.608 & 0.753 & -- & -- \\
& FRMSE mid & 0.185 & 0.859 & 0.787 & -- & -- & -- & -- & -- & -- & 0.139 & 0.565 & 0.458 & -- & -- \\
& FRMSE high & 0.045 & 0.167 & 0.128 & -- & -- & -- & -- & -- & -- & 0.012 & 0.048 & 0.049 & -- & -- \\
\hline
\multirow{4}{*}{\textbf{AGNO}} & RMSE & \textbf{0.032} & \textbf{0.166} & \textbf{0.139} & \textbf{0.116} & 0.496 & 0.413 & \textbf{0.045} & \textbf{0.354} & \textbf{0.366} & \textbf{0.007} & \textbf{0.029} & \textbf{0.030} & \textbf{0.033} & \textbf{0.014} \\
& FRMSE low & \textbf{0.101} & \textbf{0.594} & \textbf{0.637} & \textbf{0.439} & \textbf{2.870} & \textbf{1.181} & \textbf{0.171} & \textbf{2.452} & \textbf{2.680} & \textbf{0.046} & \textbf{0.207} & \textbf{0.308} & -- & -- \\
& FRMSE mid & \textbf{0.179} & 0.858 & 0.775 & \textbf{0.619} & 2.946 & 2.173 & \textbf{0.335} & 3.231 & 3.366 & \textbf{0.048} & \textbf{0.194} & \textbf{0.220} & -- & -- \\
& FRMSE high & \textbf{0.030} & 0.156 & 0.128 & 0.114 & 0.460 & 0.396 & \textbf{0.044} & 0.357 & 0.365 & \textbf{0.006} & \textbf{0.024} & \textbf{0.025} & -- & -- \\
\hline
\end{tabular}}
\caption{The main results of multi-field prediction for multiphase flow cases and multiphysics drug delivery. The lack of results is attributed to the poor training stability exhibited by the model. The Drug Delivery case is not suitable for FRMSE testing due to dataset structure issues.}
\label{frmse}
\end{table*}

\section{Experiments}
\textbf{Datasets.} 
Our experiments utilize four multiphase flow cases \cite{bubble}, which model phenomena composed of two or more phases governed by coupled physics (e.g., bubble dynamics and heat conduction). The datasets cover single bubble boiling, saturated pool boiling, subcooled pool boiling, and gravity flow boiling, with temperature $\mathbf{T}$ and velocity fields $\mathbf{V_x}$, $\mathbf{V_y}$ as prediction targets.
A respiratory drug delivery dataset \cite{peg} is further adopted to verify the generalization of AGNO. It simulates airflow-drug concentration coupling during inhalation, governed by incompressible Navier–Stokes equations and advection–diffusion conservation laws. Prediction targets are concentration $\mathbf{C}$ and pressure $\mathbf{P}$ fields.

\textbf{Baseline Models.} We compare the proposed AGNO against a set of representative baselines. For frameworks, we include MIONet \cite{mion}, U-Net \cite{unet}, GNOT \cite{GNOT}, CODA-NO \cite{coda-no}, Transolver \cite{transolver} and M2NO \cite{m2no}, covering classical DeepONet and U-Net variants, spectral representation-based NOs, and Transformer-based NOs.
For gradient balancing strategies, we include the Dual-Balancing \cite{Dual-Balancing} and ConFIG \cite{ConFIG}, both methods are integrated into the same FNO backbone, denoted as Dual-Balancing$_{\text{FNO}}$ and ConFIG$_{\text{FNO}}$.

\textbf{Evaluation Metrics.} We employ two metrics to rigorously evaluate prediction accuracy: Root Mean Square Error (RMSE) and Frequency-Dependent Root Mean Square Error (FRMSE)~\cite{bubble}.

FRMSE is utilized to test the capability in capturing physical fields across different scales, as multiphysics fields often exhibit distinct spatial frequencies. The metric is defined in the Fourier domain:
$\text{FRMSE} = \sqrt{\frac{\sum^{k_{\max}}_{k = k_{\min}}\left|\mathcal{F}(\hat{f})(k)-\mathcal{F}(f)(k)\right|^2}{k_{\max}-k_{\min}+1}},
$
where $\mathcal{F}$ denotes the discrete Fourier transformation, and $k$ represents the Fourier coordinates. Three wavenumber bands are defined to calculate FRMSE-low, -mid, -high frequency for decoupling multi-scale flow errors and quantifying the model’s performance in capturing cross-scale flow features; detailed settings are provided in the Appendix.

\subsection{Experimental Results and Discussion}

\begin{figure*}[t]
\begin{center}
\includegraphics[width=1.02\textwidth]{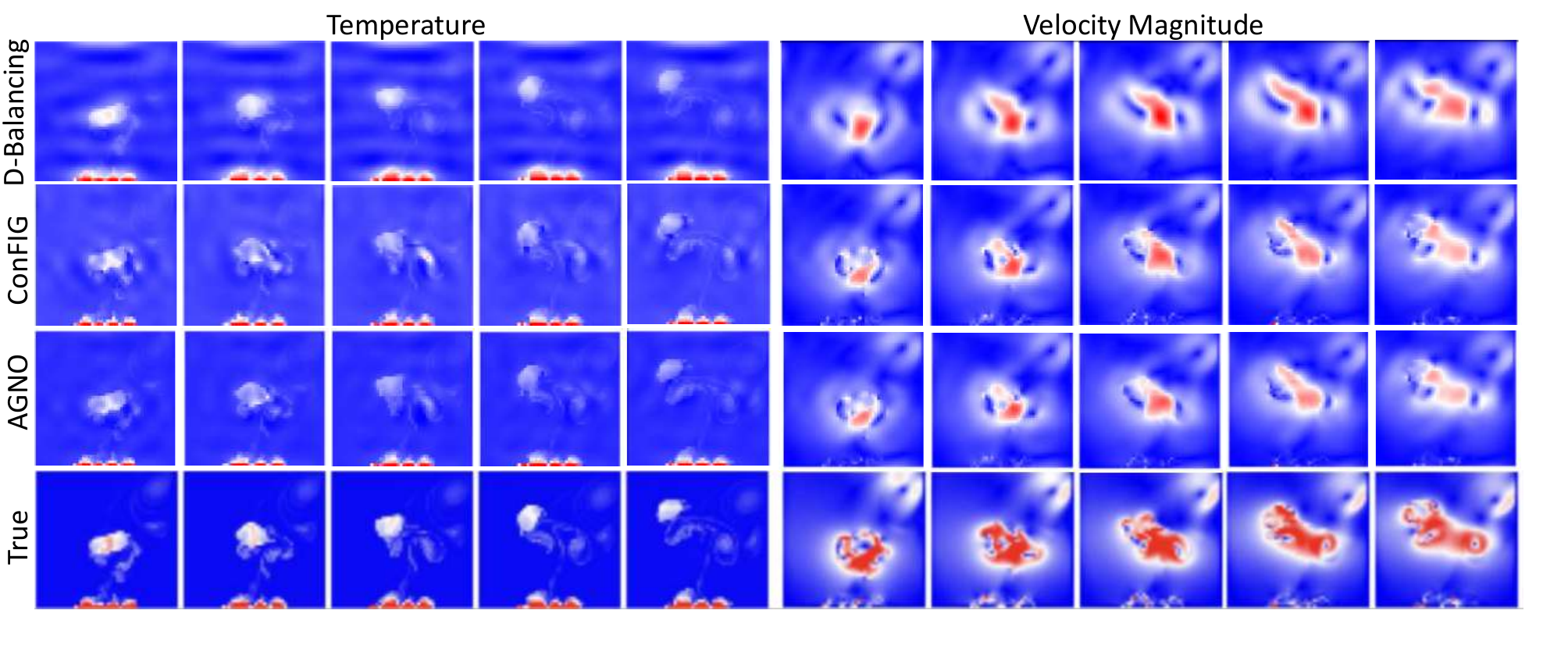}
\end{center}
\vspace{-0.4cm}
\caption{The prediction results of Dual-Balancing, ConFIG and AGNO in 41-45th time steps on the Subcooled case. Dual-Balancing is abbreviated as D-Balancing}
\label{pred}
\vspace{-0.4cm}
\end{figure*}

The performance of AGNO is evaluated against state-of-the-art baselines across five scenarios: Subcooled, Gravity, Saturated, Single Bubble and Drug Delivery. The results regarding RMSE and FRMSE are summarized in Table \ref{frmse}, respectively. The experimental results of noise robustness analysis, super-resolution performance and rollout prediction can be found in the Appendix.

\textbf{Overall Performance.} AGNO achieves the best overall performance, with the lowest RMSE in 12 of 14 cases. This confirms the effectiveness of adaptively scaling gradients from coupled fields. For example, in Subcooled, AGNO reduces $\mathbf{V_y}$ RMSE to $0.139$—about $37\%$ lower than MIONet ($0.221$). 
These results validate that AGNO balances field-specific influences during optimization, preventing dominant fields like temperature from overshadowing others. In the {Single Bubble} case, AGNO achieves RMSEs of $0.007$ ($\mathbf{T}$), $0.029$ ($\mathbf{V_x}$), and $0.030$ ($\mathbf{V_y}$), cutting velocity errors by over $50\%$ versus GNOT and U-Net. This highlights its advantage in capturing sharp, localized phenomena where small velocity inaccuracies have large impacts. In the more complex Drug Delivery case, dominated by the Navier-Stokes equations, AGNOs achieved RMSEs of $0.033$ (C) and $0.014$ (P), respectively, a significant advantage over Transolver's RMSEs of $0.120$ (C) and $0.097$ (P). This demonstrates the generalization of AGNOs. 

\textbf{Analysis in Frequency Domain.} The FRMSE results highlight AGNO's capability to resolve multi-scale features and mitigate field dominance bias. AGNO demonstrates superior performance, e.g., Subcooled $\mathbf{V_x}$ FRMSE-low improved from $1.285$ (GNOT) to ${0.594}$ (AGNO). These regimes encompass most of the field’s energy and large-scale structures. The consistent FRMSE reduction in lower-energy fields like $\mathbf{V_x}$ and $\mathbf{V_y}$ confirms that AGNO’s adaptive gradient mechanism enhances learning for subtle fields, preventing their patterns from being dominated by high-energy fields such as $\mathbf{T}$. In the high-frequency regime, which captures fine details, AGNO is competitive and often outperforms baselines. For the complex Single Bubble case, it achieves the lowest FRMSE-high across all fields, indicating that balanced learning preserves high-frequency content without sacrificing geometric feature extraction.

\textbf{Comparison with Gradient Balancing Strategies.} Although Dual-Balancing produces the lowest FRMSE mid-to-high of $\mathbf{V_x}$ and $\mathbf{V_y}$ under the subcooled, Gravity, and Saturated cases, its FRMSE low is several times larger than that of other baselines, indicating that it excessively prioritizes mid- and high-frequency signals while failing to adequately capture low-frequency components. Meanwhile, ConFIG cannot generate valid predictions owing to unstable training. These observations reveal that conventional gradient balancing methods only align gradient magnitudes or directions across distinct fields and fail to establish a unified balancing paradigm applicable to complex coupled multiphysics tasks. In contrast, our proposed AGNO framework leverages Shapley values for multiphysics decoupling and achieves consistently superior overall performance.

\begin{figure*}[t]
\begin{center}
\label{wave}
\includegraphics[width=0.91\textwidth]{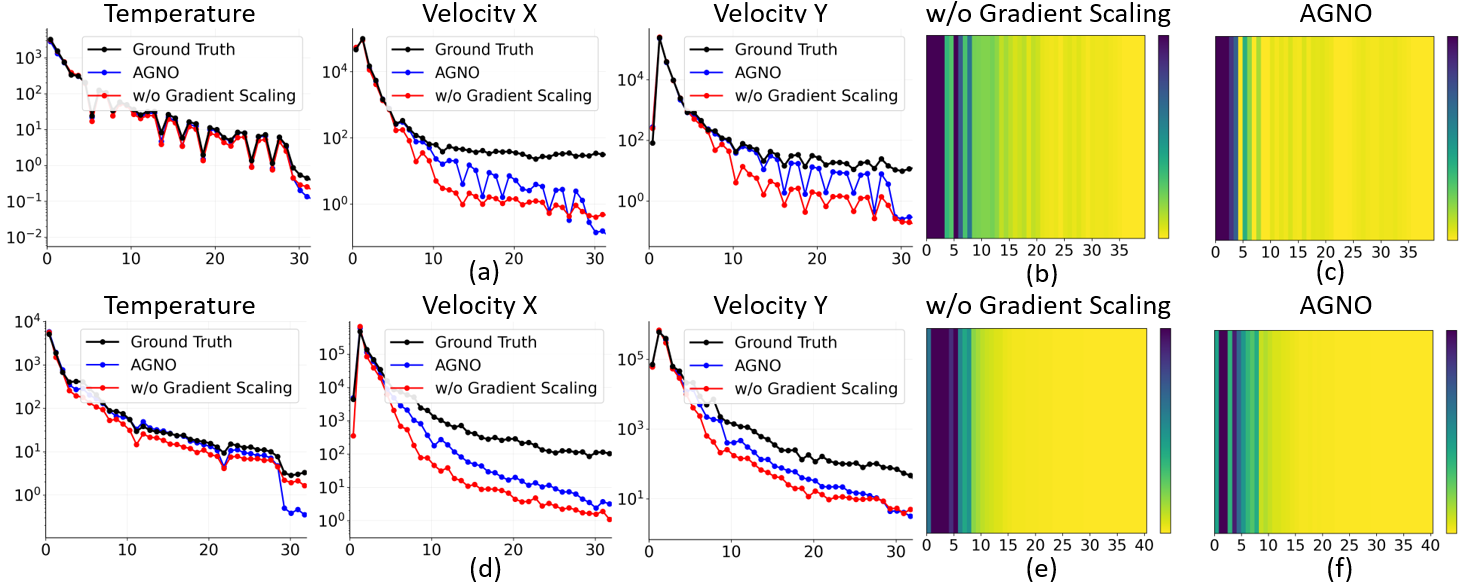}
\end{center}
\caption{The true, predicted spectral energy (a,d) and the cumulative spectral energy errors (b,c,e,f) of all fields across different wave number modes for the Subcooled and Saturated cases.}
\label{fig:spectral}
\end{figure*}

\vspace{0.1cm}
\textbf{Visualization.} Visualization results of all baseline models in the Appendix reveal that the velocity fields predicted by state-of-the-art neural operator architectures suffer from blurring and distortion, which is a typical flaw induced by field-dominant bias. As illustrated in Fig. \ref{pred}, Dual-Balancing and ConFIG introduce global prediction errors due to their non-decoupled gradient balancing strategies. In contrast, AGNO captures fine-grained details like recirculation zones and sharp wall gradients with high fidelity, which verifies the superiority of AGNO.
\vspace{-0.1cm}

\begin{table*}[htbp]\scriptsize
\centering
\begin{tabular}{ll||ccc|ccc|ccc|ccc|cc}
\specialrule{\heavyrulewidth}{0pt}{0pt}
&& \multicolumn{3}{c}{\textbf{Subcooled}} & \multicolumn{3}{c}{\textbf{Gravity}} & \multicolumn{3}{c}{\textbf{Saturated}} & \multicolumn{3}{c}{\textbf{Single Bubble}} & \multicolumn{2}{c}{\textbf{Drug Delivery}}\\
\cline{3-16}
\multirow{-2}{*}{\textbf{Model}} & \multirow{-2}{*}{\textbf{Metric}} & $\mathbf{T}$ & $\mathbf{V_x}$ & $\mathbf{V_y}$ & $\mathbf{T}$ & $\mathbf{V_x}$ & $\mathbf{V_y}$ & $\mathbf{T}$ & $\mathbf{V_x}$ & $\mathbf{V_y}$ & $\mathbf{T}$ & $\mathbf{V_x}$ & $\mathbf{V_y}$ & $\mathbf{C}$ & $\mathbf{P}$\\
\hline \hline
\multirow{3}{*}{w/o Gradient Scaling} & RMSE & 0.036 & 0.187 & 0.167 & 0.121 & 0.551 & 0.436 & 0.052 & 0.452 & 0.468 & 0.014 & 0.084 & 0.098 & 0.038 & 0.060 \\
& FRMSE low & 0.140 & 0.921 & 0.960 & 0.548 & 4.019 & 1.671 & 0.265 & 3.486 & 3.668 & \textbf{0.044} & 0.217 & 0.335 & -- & --\\
(FNO-based) & FRMSE mid & 0.200 & 0.984 & 0.940 & 0.679 & 3.398 & 2.831 & 0.376 & 3.559 & 3.734 & 0.055 & 0.265 & 0.376 & -- & --\\
& FRMSE high & 0.034 & 0.175 & 0.151 & \textbf{0.113} & 0.509 & 0.422 & 0.048 & 0.403 & 0.415 & 0.014 & 0.074 & 0.077 & -- & --\\
\hline
\multirow{3}{*}{\textbf{AGNO}} & RMSE & \textbf{0.032} & \textbf{0.166} & \textbf{0.139} & \textbf{0.116} & \textbf{0.496} & \textbf{0.413} & \textbf{0.045} & \textbf{0.354} & \textbf{0.366} & \textbf{0.007} & \textbf{0.029} & \textbf{0.030} & \textbf{0.033} & \textbf{0.014} \\
& FRMSE low & \textbf{0.101} & \textbf{0.594} & \textbf{0.637} & \textbf{0.439} & \textbf{2.870} & \textbf{1.181} & \textbf{0.171} & \textbf{2.452} & \textbf{2.680} & 0.046 & \textbf{0.207} & \textbf{0.308} & -- & --\\
(FNO-based) & FRMSE mid & \textbf{0.179} & \textbf{0.858} & \textbf{0.775} & \textbf{0.619} & \textbf{2.946} & \textbf{2.173} & \textbf{0.335} & \textbf{3.231} & \textbf{3.366} & \textbf{0.048} & \textbf{0.194} & \textbf{0.220} & -- & --\\
& FRMSE high & \textbf{0.030} & \textbf{0.156} & \textbf{0.128} & 0.114 & \textbf{0.460} & \textbf{0.396} & \textbf{0.044} & \textbf{0.357} & \textbf{0.365} & \textbf{0.006} & \textbf{0.024} & \textbf{0.025} & -- & --\\
\hline \hline
\multirow{3}{*}{w/o Gradient Scaling} & RMSE  &0.036&0.190&0.172&0.110&0.541&0.436&0.047&\textbf{0.472}&\textbf{0.484}&0.027&0.109&0.208&0.232&0.367\\
& FRMSE low &0.180&0.954&0.977& 0.500&3.893&1.784&0.214&\textbf{3.598}&\textbf{3.914}&0.152&0.503&1.027 & -- & --\\
(Transformer-based) & FRMSE mid  &0.200&0.989&0.966&0.596&3.193&2.720&0.332&\textbf{3.614}&\textbf{3.681}&0.096&0.358&0.929 & -- & --\\
& FRMSE high &0.033&0.178&0.158&0.106&0.507&0.419&0.044&\textbf{0.426}&\textbf{0.435}&0.020&0.094& 0.146 & -- & --\\
\hline
\multirow{3}{*}{\textbf{AGNO-T}} & RMSE & \textbf{0.033} & \textbf{0.186} & \textbf{0.167}&\textbf{0.102}&\textbf{0.527}&\textbf{0.434}&\textbf{0.044}&0.474&0.491&\textbf{0.009}&\textbf{0.046}&\textbf{0.050}&\textbf{0.110}&\textbf{0.200}\\
& FRMSE low & \textbf{0.138} & \textbf{0.874} & \textbf{0.985}&\textbf{0.437}&\textbf{3.650}&\textbf{1.769}&\textbf{0.190}&3.678&4.028&\textbf{0.090}&\textbf{0.544}&\textbf{0.575}& -- & --\\
(Transformer-based) & FRMSE mid & \textbf{0.179} & \textbf{0.957} & \textbf{0.931}&\textbf{0.563}&\textbf{3.135}&\textbf{2.674}&\textbf{0.302}&3.618&3.724&\textbf{0.079}&\textbf{0.424}&\textbf{0.414}& -- & --\\
& FRMSE high & \textbf{0.031} & \textbf{0.176} & \textbf{0.153}&\textbf{0.098}&\textbf{0.494}&\textbf{0.418}&\textbf{0.042}&0.427&0.440&\textbf{0.008}&\textbf{0.041}&\textbf{0.045} & -- & --\\
\hline
\end{tabular}
\caption{Ablation results for multi-field prediction, based on the FNO/Transformer architecture.}
\label{xr}
\end{table*}

\subsection{Computational Cost} Across all baselines, AGNO exhibits the lowest inference time (0.15s) and FLOPs (78M), with the exception of the two gradient balancing methods (details are provided in the Appendix). Nevertheless, AGNO struggles with prediction tasks involving a large set of physical fields. Therefore, we adopt Monte Carlo simulated Shapley values to replace (\ref{sha}) when dealing with a large number of physical fields:
\begin{equation}
\hat\phi^{m}(x)=\frac{1}{N}\sum_{i=1}^{N}V_{m\left(S_{i};\ \phi\right)}
\end{equation}
, where $N$ is the number of sampling times and $S_i$ is the subset drawn in the $i$-th random sampling that does not contain $m$.

To verify the approximation of Shapley values under realistic physical field conditions, we conducted approximation calculations across nine physical fields. The results demonstrate that our proposed method cuts computational costs by approximately 94\% while restricting the calculation error of Shapley values within 2\% (see the Appendix for full details).

\subsection{Scalability of Other Architectures} 
Our experiments also included replacing the model's backbone with ablation learning using Transformer-based NO, such as Bubbleformer~\cite{hassan2025bubbleformer}. 
The results are shown in Table \ref{xr}. Among the baselines, the Transformer-based AGNO (AGNO-T) still exhibits competitive advantages. Compared to the Transformer operator without gradient scaling, performance is improved in 12 out of 14 scenarios.
Extensive ablation results confirm that the advantages of our strategy are not restricted to a single network type, and it maintains stable superiority when combined with different mainstream neural operators.

\subsection{Ablation Study}
Ablation studies in Table \ref{xr} verify the efficacy of the proposed gradient scaling. AGNO outperforms the baseline FNO across all cases with lower overall RMSE. Improvements are particularly pronounced for low-energy, fine-grained velocity fields. In the Subcooled case, AGNO reduces $\mathbf{V_y}$ RMSE by 17\%. As shown in \ref{fig:spectral}, AGNO yields more accurate spectral energy distributions (a,d) and lower cumulative spectral errors (b,c,e,f) for both Subcooled and Saturated scenarios, enabling balanced spectral learning. These results confirm that AGNO mitigates temperature-dominated learning bias by adaptively enhancing velocity field gradients. Consistent FRMSE reductions across all frequency bands further validate its superiority in capturing fine-scale flow structures.

\section{Conclusions}

This work identified and mitigated field dominance bias in NOs for multi-variate systems, where high-energy fields dominate learning at the expense of low-energy ones. AGNO achieves balanced learning by fundamentally disentangling the backpropagated gradients corresponding to each coupled physical field and uses Shapley values to quantify each field's contribution to the total loss, enabling an adaptive gradient scaling mechanism that balances learning across all fields. Our method alleviates the bias, yielding an up to 18\% improvement in overall accuracy 
compared to the baselines.

\bibliographystyle{IEEEtran}
\bibliography{aaai2027}
\clearpage
\appendix

\section{Problem Formulation}

We consider multiphysics systems governed by $\text{PDE}$s with parameters $\theta$ defined over a spatial domain $\Omega \subset \mathbb{R}^d$ and a temporal interval $t \in [0, T]$. Let the input function space be denoted by $\mathcal{A} = \mathcal{H}_1 \times \cdots \times \mathcal{H}_m \times \theta$. Here, each $\mathcal{H}_j$ ($j = 1, \ldots, m$) represents a distinct physical field (e.g., velocity, pressure, or source terms defined over $\Omega$). The term $\theta \in \mathbb{R}^p$ denotes system parameters (e.g., viscosity ratios).

Our primary objective is to learn a neural operator $\mathcal{G}: \mathcal{A} \rightarrow \mathcal{U}$ that maps an input tuple $a = (h_1(\cdot,t), \ldots, h_m(\cdot,t))$ to the time-dependent solution field $u(\cdot, t+\Delta t) \in \mathcal{U}$. The input tuple $a$ comprises physical fields $h_i\in \mathcal{H}_i$. The output solution field $u: \Omega \times [0, T] \rightarrow \mathbb{R}^n$ is required to satisfy the governing $\text{PDE}$s. The operator $\mathcal{G}$ explicitly incorporates all input physical fields to accurately predict the evolution of the multiphysics dynamics.

\section{Related work}
NOs learn mappings between infinite-dimensional function spaces. Theoretical foundations were laid by universal approximation capabilities for operators \cite{chen1,chen2}. DeepONet \cite{lu2021learning, dyna}, the first implementation, uses dual subnetworks to encode input functions and has inspired many variants. These include MIONet \cite{mion}, which handles multiple physical fields, and physics-informed versions like PiDeepONet \cite{wang2021learning, pinns}. Subsequently, gradient balancing methods emerged to address the multi-task learning bias problem in physically-informed versions, such as Dual-Balancing and ConFIG \cite{Dual-Balancing, ConFIG}.

Another key approach uses spectral representations, exemplified by the FNO \cite{fno}. FNO leverages the  Fourier transform to parameterize integral kernels \cite{fnoqian} and has shown success in many areas \cite{FourCastNet,ace}. Enhancements like CoDA-NO \cite{coda-no} extend FNO to multiphysics fields by treating variables as tokens. Several methods, including PDE-FNO \cite{pdeFNO}, MscaleFNO \cite{MscaleFNO}, FreqMoE \cite{FreqMoE}, and DiffFNO \cite{difffno}, aim to enhance the learning of multi-scale features. Furthermore, to address the limitations of FNO in spatial frequency localization, wavelet transform-based neural operators (WNO) have emerged \cite{xb2}. WNO and its variants, such as M2NO \cite{m2no}, have achieved excellent performance in solving PDEs with local mutations, handling non-stationary signals, and modeling multi-scale physical fields.

Additionally, transformer-based neural operators including GNOT \cite{GNOT} and Transolver \cite{transolver} have recently attracted widespread attention. Featuring global modeling ability and compatibility with unstructured meshes, they can handle various complex physical problems by adopting domain-specific attention mechanisms.

\section{Pseudocode}

\subsection{Field Contribution Evaluation} It takes multiphysics fields as input to obtain the Shapley value of each physics field, which represents the contribution of each physics field to the prediction. Taking the case of three physics fields as an example, it is necessary to calculate the predicted field $f_{\text{1,2,3}}$ of the neural operator with all subsets of multiphysics fields, including the full physics fields as input. This includes the predicted fields $f_{1,3}$,$f_{2,3}$,$f_{1,2}$ with each single physics field input masked, the predicted fields $f_{{1}}$,$f_{{2}}$,$f_{{3}}$ with each pair of physics fields input masked. To avoid introducing additional information, we mask the physics field. Finally, the Shapley value of each physics field is calculated by Equation 2.

\begin{algorithm}[!h]
    \caption{Shapley Contribution Module (3 fields case)}
    \label{alg:shapley}
    
        \textbf{Input}: fields: $x_1$, $x_2$, $x_3$\\
        \textbf{Output}: Shapley values: $\{\phi^1,\phi^2,\phi^3\}$, fused output $f_{total}$
\begin{algorithmic}[1]
        \STATE $f_{total} = f(x_1, x_2, x_3)$ // All fields present
        \STATE $\mathcal{L}_{1,2,3} = MSE(f_{total}, f_{true})$

        \STATE $f_{2,3} = f(x_2, x_3)$ // Mask field 1
        \STATE $\mathcal{L}_{2,3} = MSE(f_{2,3}, f_{true})$
        \STATE $f_{1,3} = f(x_1, x_3)$ // Mask field 2
        \STATE $\mathcal{L}_{1,3} = MSE(f_{1,3}, f_{true})$
        \STATE $f_{1,2} = f(x_1, x_2)$ // Mask field 3
        \STATE $\mathcal{L}_{1,2} = MSE(f_{1,2}, f_{true})$

        \STATE $f_{{3}} = f(x_3)$ // Mask fields 1 and 2
        \STATE $\mathcal{L}_{3} = MSE(f_{3}, f_{true})$
        \STATE $f_{{2}} = f(x_2)$ // Mask fields 1 and 3
        \STATE $\mathcal{L}_{2} = MSE(f_{2}, f_{true})$
        \STATE $f_{{1}} = f(x_1)$ // Mask fields 2 and 3
        \STATE $\mathcal{L}_{1} = MSE(f_{1}, f_{true})$

        \STATE $\phi^1 = \frac{1}{3}(\mathcal{L}_{1,2,3} - \mathcal{L}_{2,3} + \mathcal{L}_{{1}}) + \frac{1}{6}(\mathcal{L}_{1,3} - \mathcal{L}_{{2}} + \mathcal{L}_{1,2} - \mathcal{L}_{{3}})$
        \STATE $\phi^2 = \frac{1}{3}(\mathcal{L}_{1,2,3} - \mathcal{L}_{1,3} + \mathcal{L}_{{2}}) + \frac{1}{6}(\mathcal{L}_{2,3} - \mathcal{L}_{{1}} + \mathcal{L}_{1,2} - \mathcal{L}_{{3}})$
        \STATE $\phi^3 = \frac{1}{3}(\mathcal{L}_{1,2,3} - \mathcal{L}_{1,2} + \mathcal{L}_{{3}}) + \frac{1}{6}(\mathcal{L}_{2,3} - \mathcal{L}_{{1}} + \mathcal{L}_{1,3} - \mathcal{L}_{{2}})$\\
        \STATE \textbf{return} $\{\phi^1,\phi^2,\phi^3\}$, $f_{total}$
    \end{algorithmic}
\end{algorithm}

\subsection{Adaptive Gradient Scaling} First, the Shapley value of each physics field and the negative MSE of the target field are used as the contribution value of each physics field. Step 2 corresponds to Equation 5, updating the contribution value $c_i^t$ of each iteration through exponential moving average. Step 3 corresponds to Equation 6, calculating the gradient scaling factors.

\begin{algorithm}[!h]
\caption{Adaptive Gradient Modulation Module}
\label{alg:grad_mod}
\textbf{Input}: Shapley values $\{\phi^1,\phi^2,\phi^3\}$, Previous EMA scores $\{c_1^{t-1}, c_2^{t-1}, c_3^{t-1}\}$\\
\textbf{Output}: Gradient scaling factors $\{\gamma_1,\gamma_2,\gamma_3\}$, Updated EMA scores $\{c_1^t, c_2^t, c_3^t\}$
\begin{algorithmic}[1]       
\IF {t = 1}
    \STATE $c_1^{0} = \phi^1$ 
    \STATE $c_2^{0} = \phi^2$
    \STATE $c_3^{0} = \phi^3$
\ENDIF
\STATE $c_1^t = 0.8 \cdot c_1^{t-1} + 0.2 \cdot \phi^1$
\STATE $c_2^t = 0.8 \cdot c_2^{t-1} + 0.2 \cdot \phi^2$
\STATE $c_3^t = 0.8 \cdot c_3^{t-1} + 0.2 \cdot \phi^3$
        
\STATE $\bar{c} = \frac{1}{3}(c_1^t + c_2^t + c_3^t)$ // Average EMA
\STATE $\gamma_1 = \exp(c_1^t - \bar{c})$
\STATE $\gamma_2 = \exp(c_2^t - \bar{c})$
\STATE $\gamma_3 = \exp(c_3^t - \bar{c})$ \\
\STATE \textbf{return} $\{\gamma_1,\gamma_2,\gamma_3\}$, $\{c_1^t, c_2^t, c_3^t\}$
\end{algorithmic}
\end{algorithm}

\section{Experiment Detail}

\subsection{Settings} All models in this study were trained using the AdamW optimizer, with a total of 50 epochs trained in the four bubble cases and 10 epochs trained in the Drug Delivery case. For hyperparameter tuning, all models supported adjustable configurations of network width (32 or 64), network depth (2 or 4 layers), and batch size (2, 4, or 8). Notably, transformer-based models were fixed with four attention heads throughout training. Slight configuration differences were adopted for different model architectures to adapt to their inherent characteristics. Specifically, the U-Net model employed a lightweight encoder to efficiently extract geometric interface information, with the encoder depth configurable as either 2 or 4 layers. Differing from U-Net, the FNO model adopted a fixed network width of 32, and provided two optional mode counts, namely 8 and 16.

Uniform downsampling strategies were implemented for dataset preprocessing. The Subcooled, Gravity, and Saturated cases were all downsampled at a factor of 8 for model training, while the Single Bubble case was downsampled at a factor of 2. For super-resolution evaluation tasks, test sets were constructed by downsampling the original datasets at factors of 4 and 2, corresponding to 4× and 2× super-resolution tasks respectively. All datasets were partitioned into training and test sets following a unified splitting protocol. The Single Bubble case contains 500 temporal snapshots, among which 450 were assigned to the training set and the remaining 50 to the test set. For the Subcooled, Gravity, Saturated, and Drug Delivery cases—each consisting of multiple simulation trajectories—one trajectory from each dataset was reserved for testing, and all remaining trajectories were used for training. Additionally, the Drug Delivery case was projected from its original spatial structure to a two-dimensional (2D) format to match the input dimensionality requirement of the models.

\subsection{Metrics Detail} 
RMSE is the primary metric for evaluating the overall accuracy of the predicted physical field across the computational domain. Its calculation is defined as:
$\text{RMSE} = \sqrt{\frac{1}{|F|}\sum_{\mathbf{x} \in F}(\hat{f}(\mathbf{x})-f(\mathbf{x}))^2}$

For FRMSE, we define three ranges for $k_{\min}$ and $k_{\max}$ to isolate errors across scales:
\begin{itemize}
    \item {FRMSE (low)}: ${[0, 4]}$, capturing large-scale flow features.
    \item {FRMSE (mid)}: ${[5, 12]}$, capturing medium-scale features.
    \item {FRMSE (high)}: ${[13, \infty]}$, representing fine-grained details and small-scale features.
\end{itemize}
These ranges quantify the error under different spectral regions, providing insight into the model's ability to capture multi-scale field details.

\subsection{Visualization} 
\begin{figure*}[!htbp]
  \centering
  \includegraphics[width=1\textwidth]{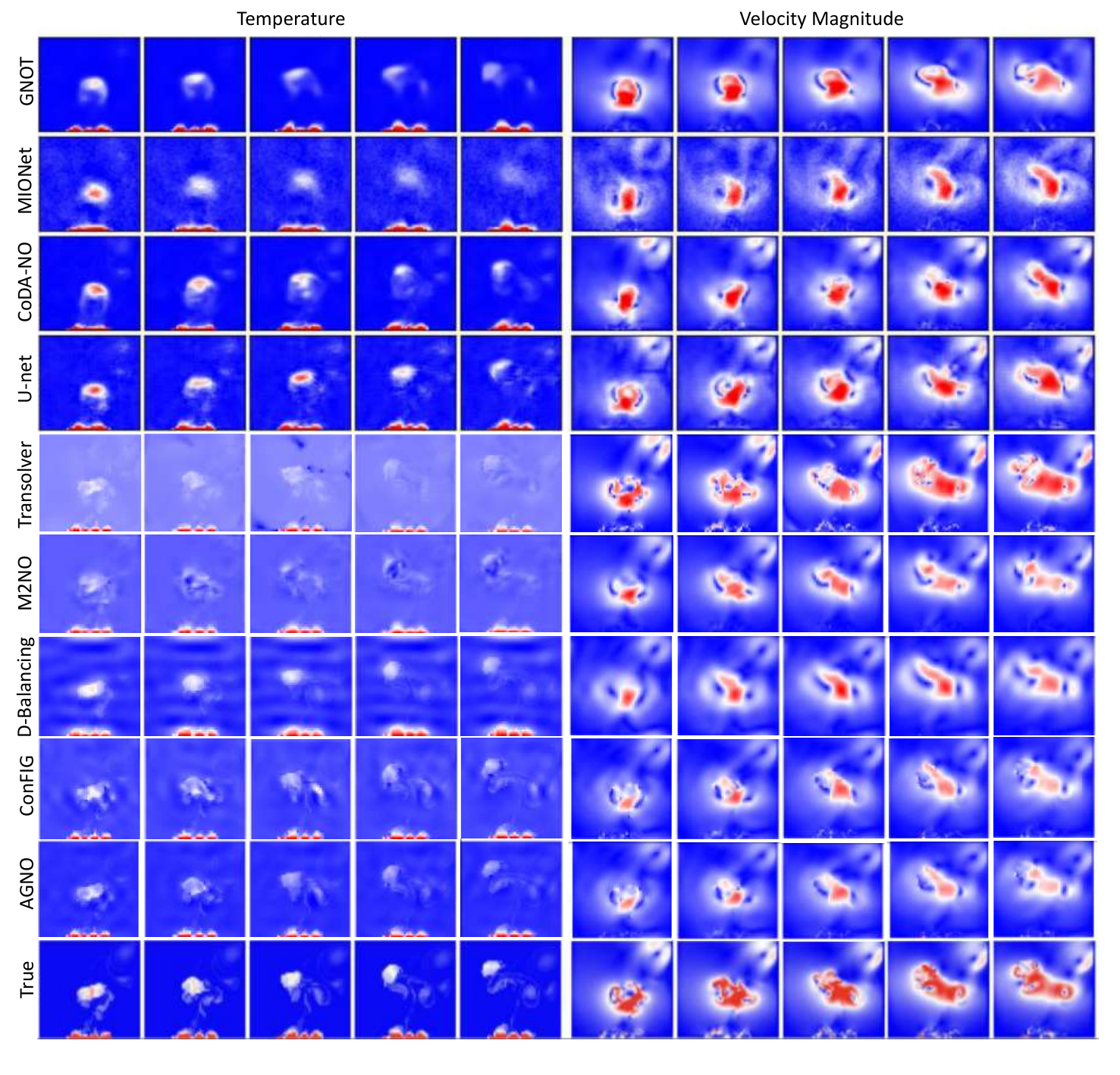}
  \caption{
  The prediction results of U-net, MIONet, CoDA-NO, GNOT, and AGNO in 41-45th time steps on the Subcooled case. Dual-Balancing is abbreviated as D-Balancing
  }
  \label{fig:subcooled_rollout1}
\end{figure*}

As shown in Fig. \ref{fig:subcooled_rollout1}, U-Net, MIONet, CoDA-NO, and GNOT produce severely accumulated prediction errors, with their velocity field contours deviating significantly from the ground truth. While Transolver and M2NO mitigate such error accumulation to a certain extent, their temperature field predictions suffer from noticeable global bias. In comparison, AGNO achieves minor cumulative errors and delivers balanced and superior prediction accuracy for both temperature and velocity fields, obtaining the optimal overall forecasting performance. 

Baseline models generate blurry velocity fields with typical field-dominated bias and suffer from progressive error accumulation at critical time steps, which further leads to the dissipation of small eddies and distorted bubble evolution. By contrast, WDNO precisely captures fine-scale flow characteristics, including recirculation zones and sharp wall gradients, and maintains outstanding spatiotemporal consistency throughout the prediction process, enabling accurate tracking of vapor phase evolution and moving contact line dynamics. These visualization results verify that the dual-branch spatial-field feature separation mechanism effectively prevents temperature signals from interfering with and masking the learning of velocity dynamics.

\subsection{Noise Robustness Analysis}
We conducted an experiment to evaluate the robustness of AGNO and baseline models under varying levels of Gaussian noise added to the training input, with results summarized in Table \ref{noise}. The Gaussian noise is defined by a standard deviation $\sigma$ calculated as $\sigma = \nu \cdot (\max(x) - \min(x))$, where $x$ represents the field points, and $\nu$ is the noise level.

\begin{table*}[!th]
\begin{center}
\resizebox{1\linewidth}{!}{
\begin{tabular}{c||cccccccccc}
\specialrule{\heavyrulewidth}{0pt}{0pt}
& MIONet & U-Net & CODA-NO & GNOT & Transolver & M2NO & Dual-Balancing$_{\text{FNO}}$ & ConFIG$_{\text{FNO}}$ & AGNO \\
\multirow{-2}{*}{Noise level} & RMSE & RMSE & RMSE & RMSE & RMSE & RMSE & RMSE & RMSE & RMSE\\
\hline \hline
1\%  &  0.478 &  0.375 &  0.412 &  0.432 & 0.401 & 0.335 & 0.923 & 0.368 &\textbf{0.336} \\
3\%   &  0.491 &  0.382 &  0.421 & 0.441 & 0.481 & 0.356 & 0.921 & 0.344 &\textbf{0.335} \\
5\%   & 0.505 & 0.395 &  0.423 & 0.445 & 0.608 & 0.392 & 0.920 & 0.344 &\textbf{0.335} \\
\hline
\end{tabular}
}
\end{center}
\label{tab:noise}
\caption{The main results of multi-field prediction for Subcooled multiphase flow cases with noisy data}
\label{noise}
\end{table*}

The results clearly demonstrate the remarkably excellent robustness of AGNO. As the noise level increases from $1\%$ to $5\%$, the RMSE of AGNO remains nearly constant, achieving the lowest error consistently (${0.336}$ at $1\%$ and ${0.335}$ at $5\%$). In contrast, the performance of most baseline models (MIONet, U-Net, CODA-NO, GNOT, Transolver, and M2NO) degrades with increasing noise, with their RMSE values rising from $1\%$ to $5\%$. For instance, Transolver's RMSE increases from $0.401$ to $0.608$, and M2NO's increases from $0.335$ to $0.392$. The results indicate that all physical fields are learned robustly, fundamentally contributing to AGNO's stable and excellent performance even when trained on noisy data.

\subsection{Super-Resolution Performance}

We evaluated the models' super-resolution generalization on a high-resolution test set, with results in Table \ref{super}. MIONet and Transolver are restricted to 1× resolution due to its structure. AGNO shows stable performance across 1×, 2×, and 4× resolutions: for temperature ($\mathbf{T}$), RMSE remains around ${0.031}$–${0.032}$; for velocity ($\mathbf{V_x}$, $\mathbf{V_y}$), RMSE stays within ${0.144}$–${0.166}$ and ${0.137}$–${0.139}$, respectively. In contrast, U-Net and CODA-NO degrade significantly—e.g., U-Net's $\mathbf{V_y}$ RMSE increases from $0.162$ (1×) to $0.267$ (4×), while CODA-NO's $T$ RMSE also increased from $0.037$ (1×) to $0.059$ (4×). This strong stability and resolution-invariant generalization demonstrate that AGNO effectively captures the underlying physical laws and multi-scale flow structures, rather than merely fitting local pixel-wise patterns, thus enabling reliable cross-resolution prediction for unsteady flow fields.

\begin{table*}[htb]
\begin{center}
\resizebox{1\linewidth}{!}{
\begin{tabular}{cc||ccccccccc}
\specialrule{\heavyrulewidth}{0pt}{0pt}
& & MIONet & U-Net & CODA-NO & GNOT & Transolver & M2NO & Dual-Balancing$_{\text{FNO}}$ & ConFIG$_{\text{FNO}}$ & AGNO \\
\multirow{-2}{*}{Fields} & \multirow{-2}{*}{Resolution} & RMSE & RMSE & RMSE & RMSE & RMSE & RMSE & RMSE & RMSE & RMSE\\
\hline \hline
$\mathbf{T}$     &1x&  0.048 &  0.036 &  0.037 &  0.038 & 0.038 & 0.034 & 0.046 & 0.037 &\textbf{0.032} \\
        &2x&  -- &      0.057 &  0.040 &  0.038 & -- & 0.037 & 0.047 & 0.038 &\textbf{0.031} \\
        &4x&  -- &      0.060 &  0.059 &  0.038 & -- & 0.048 & 0.052 & 0.037 &\textbf{0.031} \\
$\mathbf{V_x}$   &1x&  0.231 &  0.185 &  0.201 &  0.211 & 0.188 & 0.162 & 0.532 & 0.182 &\textbf{0.166} \\
        &2x&  -- &      0.213 &  0.195 &  0.199 & -- & 0.196 & 0.544 & 0.163 &\textbf{0.144} \\
        &4x&  -- &      0.228 &  0.215 &  0.194 & -- & 0.228 & 0.537 & 0.156 &\textbf{0.152} \\
$\mathbf{V_y}$   &1x&  0.221 &  0.165 &  0.178 &  0.217 & 0.162 & 0.136 & 0.345 & 0.145 &\textbf{0.139} \\
        &2x&  -- &      0.226 &  0.193 &  0.215 & -- & 0.180 & 0.344 & 0.151 &\textbf{0.137} \\
        &4x&  -- &      0.267 &  0.238 &  0.217 & -- & 0.215 & 0.347 & 0.151 &\textbf{0.138} \\
\hline
\end{tabular}}
\end{center}
\caption{The results of super-resolution evaluation on the Subcooled case. MIONet and Transolver do not support super-resolution, so results are only provided at 1x resolution.}
\label{super}
\end{table*}

\subsection{Rollout Prediction}

To validate the temporal prediction robust-ness of the proposed AGNO in capturing the dynamic evolution of multiphase flows, we conduct rollout prediction experiments in the representative Subcooled case, with specific parameter settings as follows: the rollout process initiates at timestep 40 and adopts a rollout timestep length of 5. The selection of timestep 40 as the starting point is determined by the physical characteristics of the Subcooled case: at this timestep, the flow field has transitioned from the initial transient nucleation stage to a stable dynamic cycle, where the coupling effects of bubble growth and shrinkage are fully established. This avoids the interference of unstable initial conditions and ensures that the prediction starts from a physically meaningful, steady-state dynamic baseline. The rollout timestep length of 5 is chosen to balance the coverage of key multiphase flow processes and the control of error accumulation:
a 5-timestep window is sufficient to encapsulate the complete dy-namic behavior of a typical vapor bubble as well as the associated local flow field adjustments. This length avoids the limitation of overly short timesteps and the problem of excessive error accumu-lation in long-timestep rollouts. 

The results of the rollout prediction are shown in the Table \ref{rollout}. The adaptive gradient scaling mechanism generally demonstrates superior performance, confirming the stability achieved by mitigating field dominance bias over multiple steps. In the Subcooled case, AGNO achieves the lowest RMSE across all fields, with a temperature error of 0.021 and velocity errors of 0.103 for $\mathbf{V_x}$ and 0.078 for $\mathbf{V_y}$. For the Single Bubble case, AGNO outperforms base-lines in velocity prediction, registering errors of 0.013 for $\mathbf{V_x}$ and 0.016 for $\mathbf{V_y}$, which is a key advantage for transient flow simulation. While MIONet shows a lower temperature error in the Gravity case and GNOT shows a lower temperature error in the Single Bubble case, AGNO remains highly competitive and consistently provides lower cumulative error across all four cases.
\begin{table*}[!htb]\scriptsize
\centering
\resizebox{1\linewidth}{!}{
\begin{tabular}{ll||ccc|ccc|ccc|ccc|cc}
\specialrule{\heavyrulewidth}{0pt}{0pt}
\multirow{2}{*}{\textbf{Model}}& \multirow{2}{*}{\textbf{Metric}} & \multicolumn{3}{c}{\textbf{Subcooled}} & \multicolumn{3}{c}{\textbf{Gravity}} & \multicolumn{3}{c}{\textbf{Saturated}} & \multicolumn{3}{c}{\textbf{Single Bubble}} & \multicolumn{2}{c}{\textbf{Drug Delivery}}\\
\cline{3-16}
& & $\mathbf{T}$ & $\mathbf{V_x}$ & $\mathbf{V_y}$ & $\mathbf{T}$ & $\mathbf{V_x}$ & $\mathbf{V_y}$ & $\mathbf{T}$ & $\mathbf{V_x}$ & $\mathbf{V_y}$ & $\mathbf{T}$ & $\mathbf{V_x}$ & $\mathbf{V_y}$ & $\mathbf{C}$ & $\mathbf{P}$\\
\hline \hline
\multirow{3}{*}{MIONet} & RMSE&0.036& 0.137& 0.097& \textbf{0.064} & 0.394 & 0.140 & 0.038& 0.220 & 0.191 & 0.016 &0.071&0.070 & 0.047 & 0.030\\
& FRMSE low& 0.105 & 0.414  & 0.406 & \textbf{0.336} & 4.108 & \textbf{0.827} & 0.124 & 0.968 &\textbf{1.193} &0.169 & 0.843 & 1.131 & -- & --\\
& FRMSE mid& 0.220 & 0.630 & 0.455 & \textbf{0.476} & 2.977& 0.972 & 0.227& 1.595 &1.315  & 0.144 & 0.651 & 0.512 & -- & --\\
& FRMSE high& 0.034& 0.134 & 0.093 & \textbf{0.060} & 0.339& 0.131 & 0.036 & 0.207& 0.178&0.014 &0.062& 0.060 & -- & --\\
\hline
\multirow{3}{*}{U-Net} & RMSE& 0.078 &0.432 & 0.577& 0.077 & 0.474 &0.158 & 0.042 & 0.262&0.244 & 0.047 &0.074 & 0.090 & 0.089 & 0.135\\
& FRMSE low& 0.751 &4.392 & 6.926& 0.696 & 5.277 & 1.089 & 0.128 &2.316 &2.364 & 0.590  & 0.967  & 1.507 & -- & --\\
& FRMSE mid& 0.372 &1.373 & 1.600 & 0.603 & 3.858& 1.210& 0.277 & 1.656 & 1.732 & 0.380 & 0.698 & 0.600 & -- & --\\
& FRMSE high& 0.067 & 0.378 & 0.479 & 0.069 & 0.397& 0.146 & 0.040 & 0.237 & 0.215 & 0.043&0.065 &  0.078 & -- & --\\
\hline
\multirow{3}{*}{CODA-NO} & RMSE& 0.047 & 0.149 & 0.116 & 0.076  & 0.635 & 0.148 & 0.048 & 0.416& 0.425 & -- & -- & -- & 0.147 & 0.076\\
& FRMSE low& 0.235& 0.712 & 0.836& 0.609 & 8.358  & 1.288 & 0.332  & 4.960  & 4.730  & --  &  -- & -- & -- & --\\
& FRMSE mid& 0.303& 0.685 & 0.500& 0.578 &  5.918 & 1.017 &  0.312 & 2.431 & 2.450 & -- & --  &  -- & -- & --\\
& FRMSE high& 0.043& 0.142 & 0.106& 0.069 & 0.494& 0.135 & 0.045 & 0.361 & 0.372 & --  &  -- &  -- & -- & --\\
\hline
\multirow{3}{*}{GNOT} & RMSE& 0.047 & 0.260 & 0.244 & 0.118 & 0.538 & 0.428  & 0.048 & 0.435 & 0.474 & \textbf{0.011} & 0.092 & 0.090 & 0.082 & 0.048\\
& FRMSE low& 0.265  & 1.931& 1.885& 0.572 & 4.904& 2.398 & 0.302 & 3.184  & 4.171 & \textbf{0.043}  &0.182 & 0.305 & -- & --\\
& FRMSE mid& 0.268 & 1.355 & 1.334& 0.641 & 3.406& 2.663 &  0.343 & 3.265  & 3.685 & \textbf{0.042}  &0.304 & 0.343 & -- & --\\
& FRMSE high& 0.044 & 0.235& 0.215& 0.113 & 0.489& 0.408 & 0.044 & 0.396  & 0.420& \textbf{0.009}  & 0.094 & 0.080 & -- & --\\
\hline
\multirow{3}{*}{Transolver} & RMSE& 0.083 & 0.139 & 0.099 & 0.082 & 0.651 & 0.147 & 0.099 & 0.342 & 0.402 & 0.085 & 0.175 & 0.059 & 0.236 & 0.091\\
& FRMSE low& 1.278 & 0.517 & 0.422 & 0.701 & 12.572 & 1.272 & 2.384 & 3.029 & 4.281 & 2.519 & 4.174 & 1.808 & -- & --\\
& FRMSE mid& 0.332 & 0.631 & 0.471 & 0.684 & 3.660 & 1.048 & 0.305 & 2.506 & 2.555 & 0.468 & 1.307 & 0.287 & -- & --\\
& FRMSE high& 0.054 & 0.135 & 0.095 & 0.072 & 0.360 & 0.133 & 0.047 & 0.305 & 0.352 & 0.040 & 0.123 & 0.033 & -- & --\\
\hline
\multirow{3}{*}{M2NO} & RMSE& 0.036 & 0.138 & 0.108 & 0.085 & 0.463 & 0.142 & 0.037 & 0.350 & 0.265 & 0.061 & 0.042 & 0.060 & 0.929 & 0.491\\
& FRMSE low& 0.109 & 0.387 & 0.865 & 0.741 & 5.766 & 1.301 & 0.132 & 2.945 & 2.129 & 0.214 & 0.232 & 0.350 & -- & --\\
& FRMSE mid& 0.219 & 0.631 & 0.492 & 0.841 & 3.785 & 0.815 & \textbf{0.218} & 2.605 & 1.747 & 0.226 & 0.157 & 0.231 & -- & --\\
& FRMSE high& 0.034 & 0.134 & 0.095 & 0.071 & 0.380 & 0.131 & 0.036 & 0.313 & 0.243 & 0.057 & 0.031 & 0.040 & -- & --\\
\hline
\multirow{3}{*}{Dual-Balancing$_{\text{FNO}}$} & RMSE& 0.040 & 0.530 & 0.343 & 0.801 & 0.582 & 0.129 & 0.040 & 0.535 & 0.343 & 0.062 & 0.032 & 0.029 & 0.111 & 0.097\\
& FRMSE low& 0.256 & 11.474 & 7.136 & 21.378 & 14.524 & 2.125 & 0.256 & 11.474 & 7.136 & 1.378 & 1.102 & 0.910 & -- & --\\
& FRMSE mid& 0.210 & \textbf{0.104} & \textbf{0.226} & 2.521 & 2.433 & 1.165 & 0.240 & \textbf{0.104} & \textbf{0.226} & 0.440 & 0.108 & 0.147 & -- & --\\
& FRMSE high& 0.037 & \textbf{0.035} & 0.084 & 0.240 & 0.303 & 0.112 & 0.037 & \textbf{0.035} & \textbf{0.084} & 0.048 & 0.012 & 0.014 & -- & --\\
\hline
\multirow{3}{*}{ConFIG$_{\text{FNO}}$} & RMSE& 0.503 & 0.461 & 0.285 & -- & -- & -- & -- & -- & -- & 0.903 & 0.275 & 0.171 & 0.190 & 0.104\\
& FRMSE low& 8.282 & 9.306 & 5.764 & -- & -- & -- & -- & -- & -- & 33.101 & 9.269 & 5.245 & -- & --\\
& FRMSE mid& 2.210 & 1.295 & 0.782 & -- & -- & -- & -- & -- & -- & 2.012 & 1.567 & 1.166 & -- & --\\
& FRMSE high& 0.302 & 0.140 & 0.087 & -- & -- & -- & -- & -- & -- & 0.136 & 0.104 & 0.089 & -- & --\\
\hline
\multirow{3}{*}{\textbf{AGNO }} & RMSE& \textbf{0.021} & \textbf{0.103} & \textbf{0.078} & 0.072 & \textbf{0.311}  & \textbf{0.117} & \textbf{0.037} & \textbf{0.160} & \textbf{0.151} & 0.012 & \textbf{0.013} & \textbf{0.016} & \textbf{0.044} & \textbf{0.022}\\
& FRMSE low& \textbf{0.036} & \textbf{0.276} & \textbf{0.293} & 0.442 & \textbf{1.945}  & 1.068 & \textbf{0.114} & \textbf{0.835} & 1.314 & 0.084  & \textbf{0.154}  &  \textbf{0.298} & -- & --\\
& FRMSE mid& \textbf{0.128} & 0.619 & 0.337 & 0.554 & \textbf{1.859}  &\textbf{0.931} & 0.235 & 1.040 & 0.903 & 0.071  &  \textbf{0.134} &  \textbf{0.138} & -- & --\\
& FRMSE high& \textbf{0.020}  & 0.097 & \textbf{0.076} & 0.066 & \textbf{0.295}&\textbf{0.104} & \textbf{0.036} & 0.152 & 0.140 &  0.012 & \textbf{0.011}  & \textbf{0.012}  & -- & --\\
\hline
\end{tabular}}
\caption{The main results of multi-field rollout prediction for Multiphase Flow Cases and Multiphysics Drug Delivery. The Drug Delivery case is not suitable for FRMSE testing due to dataset structure issues. The best results are shown in bold.}
\label{rollout}
\end{table*}

\subsection{Computational Cost}

\begin{table*}[!htb]
\begin{center}
\resizebox{1\linewidth}{!}{
\begin{tabular}{ccccccccccc}
\specialrule{\heavyrulewidth}{0pt}{0pt}
 & MIONet& U-Net& CODA-NO& GNOT& Transolver& M2NO& Dual-Balancing$_{\text{FNO}}$& ConFIG$_{\text{FNO}}$& AGNO \\
\hline \hline
Inference Time (s) &  1.21& 0.78 &1.52 &0.30 & 1.18 & 1.25 & 0.13 & 0.12 &0.15 \\
FLOPs (M) &  211& 484& 4855& 2593& 184 & 10966 & 10 & 10 &78\\
\hline
\end{tabular}}
\end{center}
\caption{The inference time per forward propagation and floating-point operations (FLOPs) of all models}
\label{fl}
\end{table*}

The computational cost of the Shapley value calculation adopted in AGNO rises exponentially with the growth of physical field quantities:
\begin{equation}
C_{m} = 2^m\cdot C_{1}
\end{equation}
, where $C_{m}$ is the computational cost of the Shapley value for m physical fields.
Table \ref{fl} shows the inference Time and FLOPs of all baselines, AGNO exhibits the lowest inference time and computational FLOPs, with the exception of the two gradient balancing methods sharing the same backbone architecture.
As shown in the Table \ref{mt}, the shapley value calculated using Monte Carlo differs from the exact value by only 2\%.
Table \ref{c2} shows the inference Time and FLOPs of Precise shapley and Monte Carlo shapley computation.

\begin{table*}[htb]
\begin{center}
\resizebox{1\linewidth}{!}{
\begin{tabular}{cccccccccc}
\specialrule{\heavyrulewidth}{0pt}{0pt}
     & Field 1& Field 2& Field 3& Field 4& Field 5& Field 6& Field 7& Field 8& Field 9 \\
\hline \hline
Precise shapley &$4.72\times10^{-4}$&$-1.02\times10^{-4}$&$-3.13\times10^{-4}$&$-3.65\times10^{-4}$&$-4.22\times10^{-4}$&$-1.56\times10^{-3}$&$-2.46\times10^{-4}$&$-3.51\times10^{-4}$&$3.43\times10^{-4}$ \\
Monte Carlo shapley &  $4.73\times10^{-4}$ & $-9.88\times10^{-5}$ & $-3.12\times10^{-4}$ & $-3.65\times10^{-4}$ & $-4.20\times10^{-4}$ & $-1.55\times10^{-3}$ & $-2.43\times10^{-4}$ & $-3.54\times10^{-4}$ & $3.41\times10^{-4}$ \\
\hline
\end{tabular}}
\end{center}
\caption{Shapley values under Monte Carlo simulations with 9 physical fields}
\label{mt}
\end{table*}

\begin{table}[!htbp]
\begin{center}
\resizebox{1\linewidth}{!}{
\begin{tabular}{cccccccccc}
\specialrule{\heavyrulewidth}{0pt}{0pt}
     & Inference Time (s)& FLOPs (M) \\
\hline \hline
Precise shapley &0.16&5003\\
Monte Carlo shapley &  0.15 & 313\\
\hline
\end{tabular}}
\end{center}
\caption{The inference time per forward propagation and floating-point operations (FLOPs) of Monte Carlo simulation}
\label{c2}
\end{table}

\section{Dataset}
\textbf{BubbleML} Developed explicitly for machine learning (ML) research, BubbleML is a comprehensive multiphysics dataset focused on phase-change boiling. Its primary objective is to alleviate the dearth of high-caliber datasets that hinder ML-based investigations into phase-change processes. A key limitation of existing experimental datasets lies in their restricted applicability and insufficient ground truth annotations; meanwhile, most contemporary scientific machine learning (SciML) datasets are confined to single-phase physical phenomena, lacking the ability to encapsulate the complex features inherent to multi-phase phase changes.
Leveraging physics-informed simulation techniques via the Flash-X framework, the dataset incorporates 79 distinct sets of simulation results, encompassing a wide spectrum of boiling configurations—nucleate pool boiling, flow boiling, and subcooled boiling among them. It covers an expansive parameter space, ranging from terrestrial to microgravity gravitational environments, alongside varied flow velocities, subcooling degrees, and wall superheat magnitudes. Essential physical quantities, including fluid velocity, temperature distribution, pressure fields, and the signed distance function that delineates liquid-vapor interfaces, are archived in the HDF5 format. To ensure its reliability and precision, the dataset has been subjected to rigorous experimental validation procedures.

To facilitate its application in ML research, two distinct types of benchmark tasks have been constructed for this dataset: one focusing on optical flow analysis to characterize bubble dynamics, and the other centered on SciML-based modeling for thermal dynamic predictions. Aligned with the FAIR (Findable, Accessible, Interoperable, Reusable) guidelines, the dataset is freely accessible to the research community via GitHub and Zenodo. This initiative addresses the existing gap in ML datasets for multi-phase phase-change studies, providing researchers with high-fidelity, diverse data resources and standardized evaluation benchmarks to advance ML-driven explorations of phase-change phenomena.

For the experimental validation portion of the study, four specific cases were selected: Subcooled Pool Boiling, Flow Boiling under Gravitational Conditions, Saturated Pool Boiling, and Single Bubble experiments.

\begin{figure*}[t]
    \centering
    \includegraphics[width=0.6\textwidth]{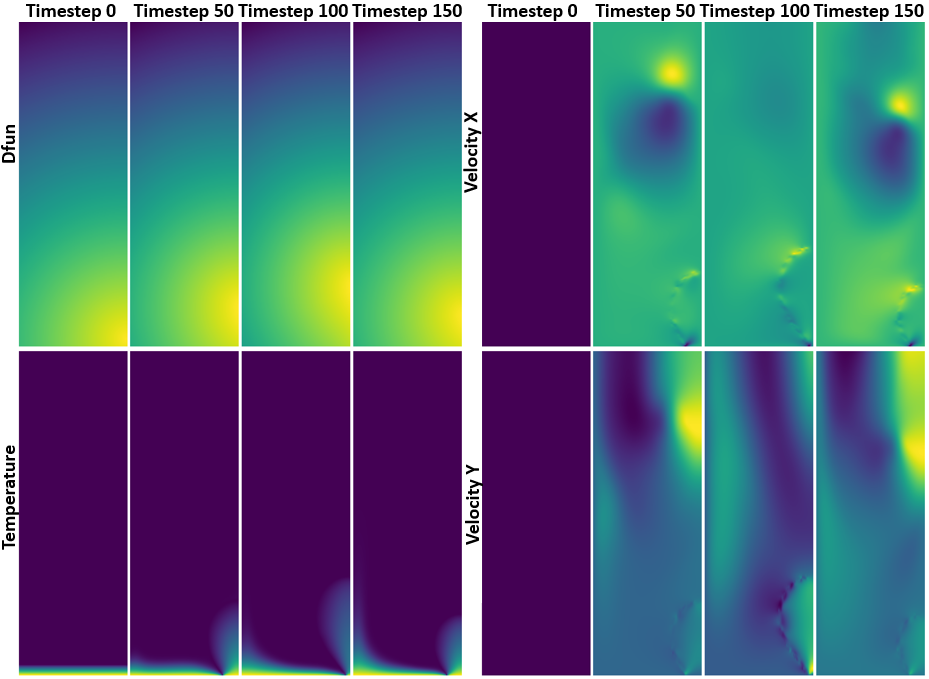}
    \caption{
    Single Bubble Visualization
    }
    \label{Single Bubble}
\end{figure*}

\textbf{Single Bubble} Within a confined spatial environment, an isolated bubble undergoes the entire sequence of its life cycle—beginning with nucleation and subsequent growth, followed by detachment from the heating element located at the bottom, and concluding with upward migration—without being affected by the presence of other bubbles. As the fundamental building block for exploring complex boiling mechanisms, this single-bubble system plays a crucial role in two key aspects: first, it facilitates the extraction of essential multiphase flow regulations, including the associations between bubble detachment rate, ascending speed, and heat input; second, it provides a basis for the adjustment and validation of machine learning models, such as simplified tasks involving the dynamic tracking of bubble movement. Below is a comprehensive breakdown of the dataset’s specifications and core configuration parameters.
The simulation scenario is composed of a single 2D model (designated as PB-Single Bubble). Key specifications of this model include a physical domain measuring 4.2 mm × 6.3 mm, a spatial resolution of 192 × 288 pixels, and a time increment of $\Delta t = 0.5$—this time step translates to a physical duration of 0.004 seconds. To ensure complete coverage of the single bubble’s life cycle, from initial nucleation to final ascent, the simulation incorporates a total of 500 time steps, resulting in an overall data volume of 0.5 GB.

All core parameters of the simulation are set uniformly, as detailed below: The working fluid selected for the simulation is FC-72 (perfluorohexane), a substance characterized by its electrical insulating properties and suitability for low-temperature cooling operations in the 50–100 °C range. Gravitational conditions are standardized to terrestrial gravity, with a/g = 1. The temperature of the heating surface is calibrated to lie within the nucleate boiling range specific to FC-72; this configuration guarantees the stable formation of bubbles while preventing any disruptions caused by condensation or bubble coalescence.

The shape of the bubble and its interface are precisely quantified using the signed distance function (SDF), represented by $\phi$. In this framework, a positive value of $\phi$ ($\phi \textgreater 0$) denotes the vapor phase, whereas a negative value ($\phi \textless 0$) corresponds to the liquid phase. The initial bubble is placed at a predetermined nucleation point on the heater surface, taking the form of a small sphere with an initial velocity of 0. Notably, the simulation is free from initial flow disturbances or pressure gradient effects, which allows for the unobstructed capture of the intrinsic physical dynamics governing the behavior of a single bubble.

\begin{figure*}[t]
    \centering
    \includegraphics[width=0.6\textwidth]{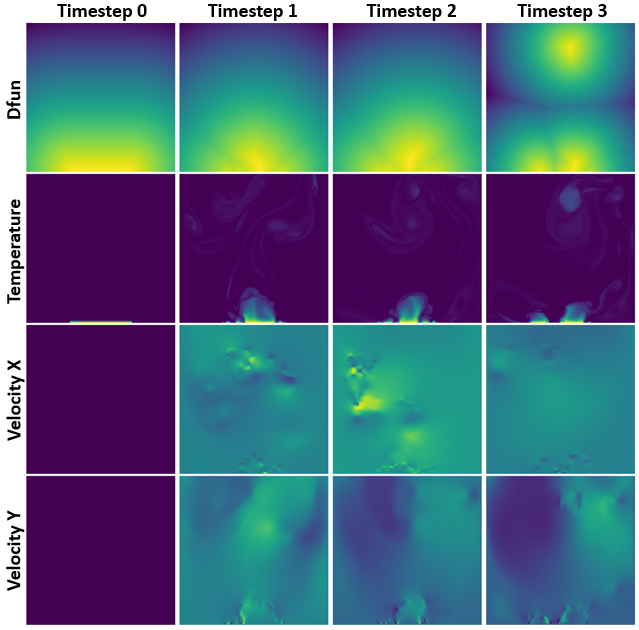}
    \caption{
    Subcooled Pool Boiling
    }
    \label{Subcooled Pool}
\end{figure*}

\textbf{Subcooled Pool Boiling} A defining feature of subcooled boiling lies in the fact that the overall temperature of the liquid phase remains lower than its saturation temperature. When bubbles initiate (nucleate) and expand on the surface of the bottom-mounted heater, they undergo condensation upon contact with the surrounding subcooled liquid—this process demonstrates dynamic behaviors characterized by the simultaneous occurrence of evaporation and condensation. This particular boiling phenomenon is mainly applied in practical scenarios that demand efficient bubble-driven heat transfer, such as the high-performance cooling of electronic components and low-temperature thermal management systems.
The detailed specifications and key parameters of the corresponding dataset are outlined as follows:

- PB-Subcooled: This subset comprises 10 independent simulation runs, with a physical computational domain measuring 8.4 mm × 8.4 mm and a spatial resolution of 384 × 384 pixels. The time step is set to $\Delta t = 1$, which corresponds to a physical time interval of 0.008 seconds, and the total number of time steps reaches 200, resulting in a total data volume of 10.3 GB.

- PB-Subcooled 0.1: Consisting of 15 simulation trials, this subset shares the same physical domain dimensions and spatial resolution as the PB-Subcooled dataset. However, it adopts a smaller time step of $\Delta t = 0.1$ (equivalent to a physical time of 0.0008 seconds) and includes 2000 time steps in total, leading to a larger data volume of 155.1 GB.

The core parameters for all simulations are uniformly configured as follows: The working fluid employed is FC-72 (perfluorohexane), a substance with excellent electrical insulation capabilities, making it well-suited for low-temperature cooling applications within the range of 50–100 °C. The subcooling effect is realized by regulating the temperature difference between the bulk liquid temperature ($\text{T}_{bulk}$) and the saturation temperature of FC-72 (around 58 °C), which ensures a distinct condensation effect during the boiling process. The heater temperature is adjusted between 60–110 °C, a range that covers the nucleate boiling regime and prevents the occurrence of critical heat flux (CHF). Gravitational conditions are set to the standard terrestrial gravity (a/g = 1) by default. The liquid-vapor interface is monitored and tracked using the Signed Distance Function (SDF), denoted as $\phi$—where $\phi > 0$ indicates the vapor phase and $\phi < 0$ denotes the liquid phase. Additionally, the boundary conditions of the simulation domain are fixed to eliminate any extraneous disturbances, thereby focusing the analysis on the impacts of subcooling degree on bubble dynamics and heat transfer flux.

\begin{figure*}[t]
    \centering
    \includegraphics[width=0.6\textwidth]{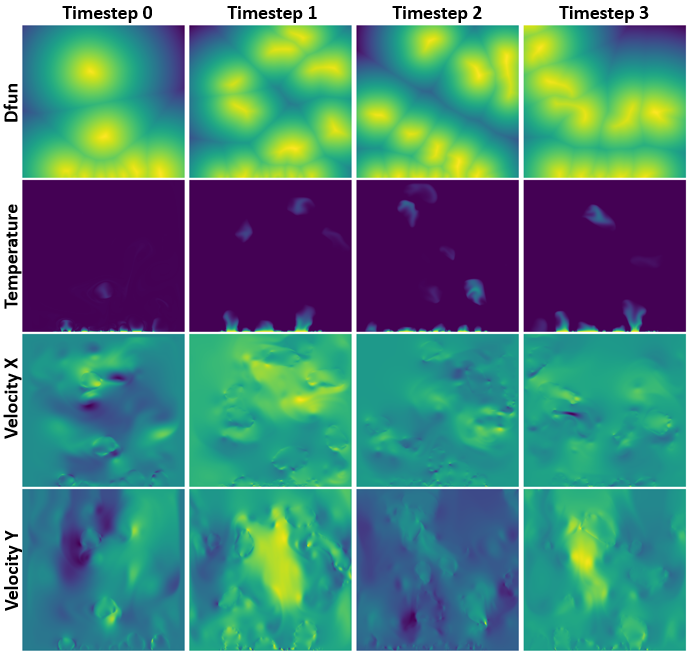}
    \caption{
    Saturated Pool Boiling
    }
    \label{Saturated Pool}
\end{figure*}

\textbf{Saturated Pool Boiling} A defining physical trait of saturated pool boiling lies in the equality between the liquid’s bulk temperature and its saturation temperature, a state where the liquid and vapor phases maintain thermal equilibrium. Upon nucleation and growth on the heater surface, bubbles do not experience condensation-driven contraction; instead, their detachment and upward movement are solely driven by buoyant forces. This particular scenario is pivotal to investigating the shift from nucleate boiling to critical heat flux, and it holds broad significance for addressing boiling heat transfer challenges in non-subcooled environments—such as those encountered in nuclear waste thermal management and electronic component heat dissipation.
The saturated pool boiling process is examined through 13 distinct, independent simulations. Heater temperatures are uniformly spaced at 5°C increments, ranging from 60°C to 120°C, which encompasses the transition of boiling flow patterns from bubbly flow to slug flow. The simulated physical domain measures 11.2mm × 11.2mm, with a spatial resolution of 512 × 512 pixels. For temporal discretization, a time step $\Delta t = 1$ (dimensionless time, equivalent to 0.008 seconds of physical time) is adopted, and each simulation incorporates 200 time steps—an interval sufficient to capture the full lifecycle of multiple bubble nucleation-detachment cycles. The total dataset amounts to 24.2GB, all of which is stored in the HDF5 file format.

FC-72 (perfluorohexane) serves as the working fluid in these simulations. This fluid exhibits electrical insulation capabilities, making it well-suited for low-temperature cooling applications within the 50–100°C range. Simulations are conducted under terrestrial gravity conditions (no low-gravity configurations are implemented), with a primary focus on capturing the fundamental physical principles governing saturated boiling in standard environments.

\begin{figure*}[t]
    \centering
    \includegraphics[width=\textwidth]{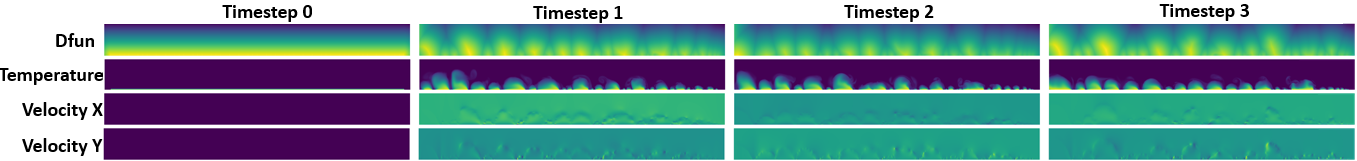}
    \caption{
    Flow Boiling Gravity
    }
    \label{Flow Boiling}
\end{figure*}

\textbf{Flow Boiling Gravity} The Flow Boiling Gravity experimental system is engineered to investigate how gravitational forces influence bubble dynamic behaviors, the evolutionary characteristics of flow regimes, and heat transfer flux. This is accomplished by adjusting gravitational environments, which span from standard terrestrial gravity conditions ($a/g=1$) to the microgravity environment encountered aboard the International Space Station ($a/g=10^{-4}$). Its primary applications are focused on practical fields such as space exploration missions and thermal management systems operating under extreme environmental conditions. Below is a comprehensive breakdown of the specifications and key parameters associated with its dataset.

- FB-Gravity Dataset: Comprising 6 independent simulation runs, this dataset features a physical computational domain measuring 35 mm × 3.5 mm, with a spatial resolution of 1600 × 160 pixels. The simulation employs a time increment of $\Delta t=1$, which translates to an actual physical time interval of 0.008 seconds. Each simulation consists of 200 discrete time steps, resulting in a total data volume of 10.9 GB.

- FB-Gravity 0.1 Dataset: This dataset also includes 6 simulation runs, maintaining the same physical domain dimensions and spatial resolution as the FB-Gravity dataset. However, it utilizes a smaller time increment of $\Delta t=0.1$ (corresponding to a physical time of 0.0008 seconds) and a greater number of time steps (2000 in total), leading to a larger overall data volume of 108.6 GB.

A set of consistent core parameters is applied across all simulations: The working fluid selected is FC-72 (perfluorohexane), a substance characterized by its electrical insulation capabilities and suitability for low-temperature cooling applications within the range of 50–100 $^\circ C$. Gravitational scaling is implemented through the Froude number, with $\text{Fr}^{-2}=a/g$, encompassing a gradient from $10^{-4}$ to 1. The inlet flow velocity is fixed at the threshold velocity required to initiate bubbly flow during flow boiling, while the heater temperature is controlled within the nucleate boiling range of FC-72 (80–100 $^\circ C$). This parameter configuration ensures stable bubble nucleation while preventing the occurrence of critical heat flux, which would disrupt the experimental conditions.

\textbf{Multiphysics Drug Delivery} The respiratory drug delivery simulation dataset is constructed to quantitatively investigate the coupled multiphysics evolution of respiratory airflow and aerosol drug concentration transport inside human upper airway anatomical structures. This dataset characterizes airflow velocity, pressure distribution and scalar drug advection-diffusion transport under varied inhalation boundary conditions and initial drug concentration loads, supporting the development of physics-embedded graph network simulators for medical respiratory thermal-fluid analysis. Its primary applications cover pulmonary drug inhalation therapy simulation, personalized respiratory treatment optimization, and unstructured mesh-based multiphysics long-term prediction research. Below is a comprehensive breakdown of the specifications and key parameters associated with its dataset.

\begin{figure*}[t]
    \centering
    \includegraphics[width=1\textwidth]{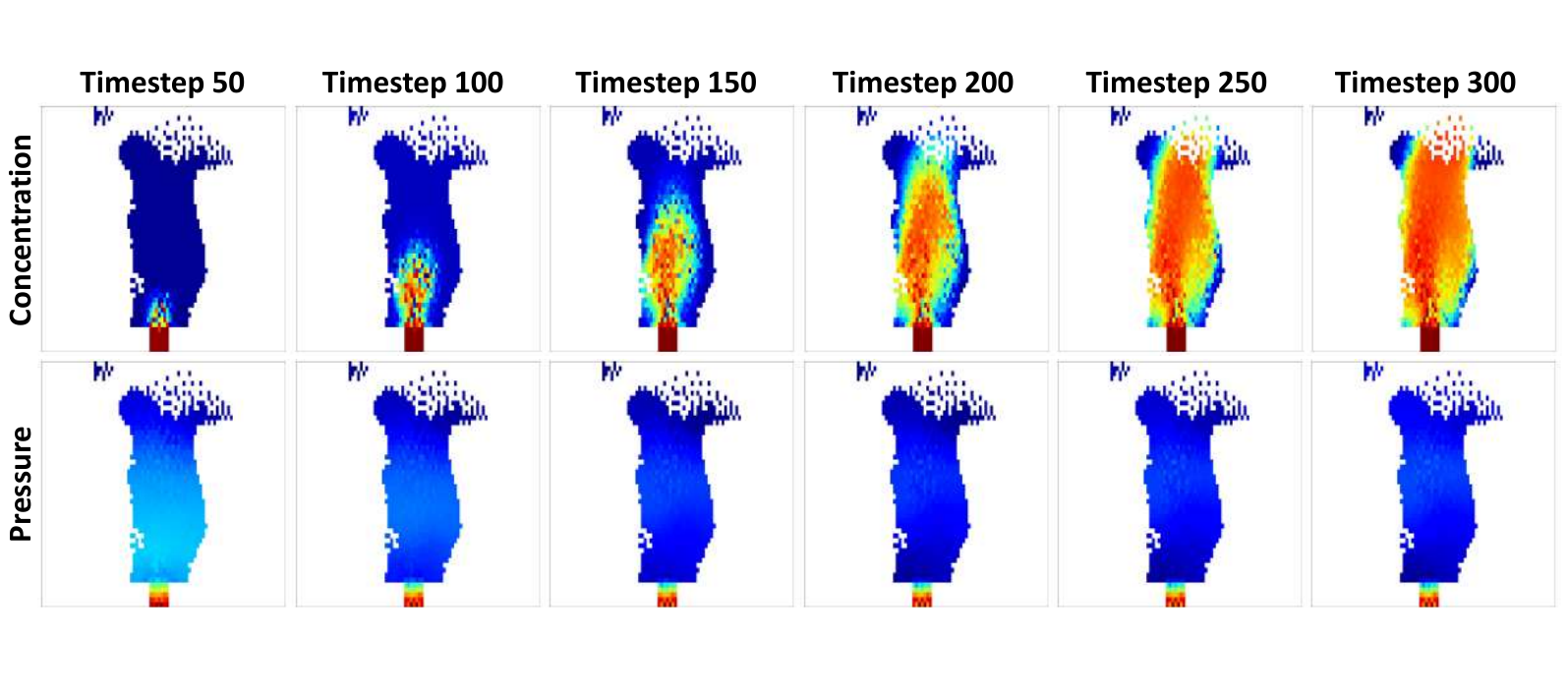}
    \caption{
    Drug Delivery
    }
    \label{drug}
\end{figure*}

Respiratory Drug Delivery Dataset: Comprising 144 independent simulation trajectories, this dataset features a 3D tetrahedral mesh computational domain reconstructed from simplified human upper respiratory tract geometry equipped with an oral drug delivery mouthpiece. The simulation employs a time increment of $\Delta t=0.005$, which corresponds to an actual physical time interval of 0.005 seconds. Each simulation trajectory consists of 400 discrete time steps, covering a complete inhalation cycle of 2 seconds, with 120 trajectories allocated for model training and 24 for testing.
A set of consistent core parameters is applied across all simulation trajectories: The governing physical laws are the incompressible Navier-Stokes equations for fluid airflow and advection-diffusion equations for scalar drug concentration transport. The spatial mesh adopts 3D tetrahedron unstructured grids to fit complex anatomical airway geometry, with node features including spatial coordinates, node type one-hot encoding, flow velocity, fluid pressure, and drug concentration scalar field. Inlet airflow boundary conditions and initial aerosol drug concentration distributions are set to multiple gradient levels to diversify physical flow states. The simulation strictly follows mass conservation and fluid divergence-free physical constraints, avoiding flow separation artifacts and unphysical mass loss, which provides high-fidelity ground truth data for evaluating the physical consistency and long-term prediction stability of multiphysics deep learning simulators.

\end{document}